# Identifying nonlinear dynamical systems via generative recurrent neural networks with applications to fMRI

Georgia Koppe[1,2], Hazem Toutounji[1,6], Peter Kirsch[3], Stefanie Lis[4], Daniel Durstewitz[1,5]

[1]Department of Theoretical Neuroscience
[2]Department of Psychiatry and Psychotherapy
[3]Department of Clinical Psychology
[4]Institute for Psychiatric and Psychosomatic Psychotherapy
Central Institute of Mental Health
Medical Faculty Mannheim, Heidelberg University
[5]Faculty of Physics and Astronomy, Heidelberg University
68159 Mannheim, Germany
[6]Institute of Neuroinformatics, University of Zurich and ETH Zurich, 8057 Zurich, Switzerland
georgia.koppe@zi-mannheim.de

**Abstract**

A major tenet in theoretical neuroscience is that cognitive and behavioral processes are ultimately implemented in terms of the neural system dynamics. Accordingly, a major aim for the analysis of neurophysiological measurements should lie in the identification of the computational dynamics underlying task processing. Here we advance a state space model (SSM) based on generative piecewise-linear recurrent neural networks (PLRNN) to assess dynamics from neuroimaging data. In contrast to many other nonlinear time series models which have been proposed for reconstructing latent dynamics, our model is easily interpretable in neural terms, amenable to systematic dynamical systems analysis of the resulting set of equations, and can straightforwardly be transformed into an equivalent continuous-time dynamical system. The major contributions of this paper are the introduction of a new observation model suitable for functional magnetic resonance imaging (fMRI) coupled to the latent PLRNN, an efficient stepwise training procedure that forces the latent model to capture the 'true' underlying dynamics rather than just fitting (or predicting) the observations, and of an empirical measure based on the Kullback-Leibler divergence to evaluate from empirical time series how well this goal of approximating the underlying dynamics has been achieved. We validate and illustrate the power of our approach on simulated 'ground-truth' dynamical systems as well as on experimental fMRI time series, and demonstrate that the learnt dynamics harbors task-related nonlinear structure that a linear dynamical model fails to capture. Given that fMRI is one of the most common techniques for measuring brain activity non-invasively in human subjects, this approach may provide a novel step toward analyzing aberrant (nonlinear) dynamics for clinical assessment or neuroscientific research.

**Author Summary**

Computational processes in the brain are often assumed to be implemented in terms of nonlinear neural network dynamics. However, experimentally we usually do not have direct access to this underlying dynamical process that generated the observed time series, but have to infer it from a sample of noisy and mixed measurements like fMRI data. Here we combine a dynamically universal recurrent neural network (RNN) model for approximating the unknown system dynamics with an observation model that links this dynamics to experimental measurements, taking fMRI data as an example. We develop a new stepwise optimization algorithm, within the statistical framework of state space models, that forces the latent RNN model toward the true data-generating dynamical process, and demonstrate its power on benchmark systems like the chaotic Lorenz attractor. We also introduce a novel, fast-to-compute measure for assessing how well this worked out in any empirical situation for which the ground truth dynamical system is not known. RNN models trained on human fMRI data this way can generate new data with the same temporal structure and properties, and exhibit interesting nonlinear dynamical phenomena related to experimental task conditions and behavioral performance. This approach can easily be generalized to many other recording modalities.

**Introduction**

A central tenet in computational neuroscience is that computational processes in the brain are ultimately implemented through (stochastic) nonlinear neural system dynamics [1-3]. This idea reaches from Hopfield's [4] early proposal on memory patterns as fixed point attractors in recurrent neural networks, working memory as rate attractors [5, 6], decision making as stochastic transitions among competing attractor states [7], motor or thought sequences as limit cycles or heteroclinic chains of saddle nodes [8, 9], to the role of line attractors in parametric working memory [10-12], neural integration [13], interval timing [14], and more recent thoughts on the role of *transient* dynamics in cognitive processing [15]. To test and further develop such theories, methods for directly assessing system dynamics from neural measurements would be of great value.

Traditionally, mostly linear approaches like linear (Gaussian or Gaussian-Poisson) state space models [16-19], Gaussian Process Factor Analysis [GPFA; 20], Dynamic Causal Modeling [DCM; 21], or (nonlinear, but model-free) delay embedding techniques [22, 23], have been used for reconstructing state space trajectories from *experimental* recordings. While these are powerful visualization tools that may also give some insight into system parameters, like connectivity [21], linear dynamical systems (DS) are *inherently* very limited with regards to the range of dynamical phenomena they can produce [e.g. 24]. The representation of smoothed trajectories in the latent space may still inform the researcher about interesting aspects of the dynamics, but the inferred latent model on its own is not powerful enough to reproduce many interesting and computationally important phenomena like multi-stability, complex limit cycles, or chaos [24, 25]. More formally, given experimental observations $\mathbf{X} = \{\mathbf{x}_t\}$ supposedly generated by some underlying latent dynamical process $\mathbf{Z} = \{\mathbf{z}_t\}$ (Fig 1), linear state space models may yield a useful approximation to the posterior $p(\mathbf{Z}|\mathbf{X})$, but – due to their linear limitations – they will not produce an adequate mathematical model of the prior dynamics $p(\mathbf{Z})$ that could generate the actual observations via $p(\mathbf{X}|\mathbf{Z})$.

In contrast, recurrent neural networks (RNNs) represent a class of nonlinear DS models which are universal in the sense that they can approximate arbitrarily closely the flow of any other dynamical system [26-28]. Hence, RNNs are, in theory, sufficiently powerful to emulate any type of brain dynamics. Based on previous work embedding RNNs into a statistical inference framework [29, 30], we have recently developed a nonlinear state space model utilizing piecewise-linear RNNs (PLRNNs) for the latent dynamical process [31]. In state space models, similar to sequential variational auto-encoders (VAEs) [32-34], one attempts to infer the system parameters $\boldsymbol{\theta}$ by maximizing a lower bound on the log-likelihood $\log p(\mathbf{X}|\boldsymbol{\theta})$. In contrast to many other RNN-based approaches [30, 35], including sequential VAEs [36], our method returns a set of neuronally interpretable and partly analytically tractable dynamical equations that could be used to gain further insight into the generating system.

The present work further advances this powerful methodology along three major directions: First, we develop a stepwise initialization and training scheme that forces the latent PLRNN model toward the correct underlying dynamics: Good prediction of the time series observations and informative smooth latent trajectories may be achieved even without recreating a sufficiently good approximation to the underlying DS (as evidenced by the success of linear state space models). Through a kind of annealing protocol that places increasingly more burden of predicting the observations onto the latent process model, we enforce the correct dynamics. Second, we show that a Kullback-Leibler divergence defined across state space between the prior generative model dynamics $p(\mathbf{Z})$ (independent of the observations) and the inferred latent states given the observations, $p(\mathbf{Z}|\mathbf{X})$, provides a good measure for how well the reconstructed DS (emulated by the PLRNN) can be expected to have captured the correct underlying system. Hence, our approach, rather than just inferring the latent space underlying the observations, attempts to force the system to capture the correct dynamics in its governing equations, and provides a quantitative sense of how well this worked for any empirically observed system for which the ground truth is not known. Third, given that fMRI is likely the most important non-invasive technique for gaining insight into human brain function in healthy subjects and psychiatric illness, we provide an observation ('decoder') model for the PLRNN that takes the hemodynamic response filtering into account.

**Results**

**PLRNN-based state space model (PLRNN-SSM)**

We start by introducing our nonlinear state space model (SSM) and statistical inference framework [originally developed in 31]. Within a SSM, one aims to predict observed experimental time series $\mathbf{x}_t \in \mathbb{R}^N$ from a potentially much lower dimensional set of latent variables $\mathbf{z}_t \in \mathbb{R}^M$ and their temporal dynamics. Here we use a piecewise-linear (or, strictly, piecewise-affine) recurrent neural network (PLRNN) (i.e., a RNN composed of rectified-linear units [ReLUs]) for modeling the unknown latent dynamics:

$$(1) \quad \mathbf{z}_t = \mathbf{A}\mathbf{z}_{t-1} + \mathbf{W}\varphi(\mathbf{z}_{t-1}) + \mathbf{h} + \mathbf{C}\mathbf{s}_t + \boldsymbol{\varepsilon}_t, \quad \boldsymbol{\varepsilon}_t \sim N(\mathbf{0}, \boldsymbol{\Sigma})$$

$$\mathbf{z}_1 \sim N(\boldsymbol{\mu}_0 + \mathbf{C}\mathbf{s}_1, \boldsymbol{\Sigma})$$

where $\mathbf{z}_t$ is the latent state vector at time $t=1...T$, $\mathbf{A} \in \mathbb{R}^{M \times M}$ is a diagonal matrix of (linear) auto-regression weights, $\mathbf{W} \in \mathbb{R}^{M \times M}$ an off-diagonal matrix of connection weights, and $\varphi(\mathbf{z}_t) = \max(\mathbf{z}_t, 0)$ is an (element-wise) ReLU transfer function. $\mathbf{s}_t \in \mathbb{R}^K$ denotes time-dependent external inputs that influence latent states through coefficient matrix $\mathbf{C} \in \mathbb{R}^{M \times K}$, and $\boldsymbol{\varepsilon}_t$ is a Gaussian white noise process with diagonal covariance matrix $\boldsymbol{\Sigma}$. (The basic model was modified from Durstewitz (31) to enable efficient estimation of bias parameters $\mathbf{h}$ and speeding up the inference algorithm by orders of magnitude.) The diagonal and off-diagonal structure of $\mathbf{A}$ and $\mathbf{W}$, respectively, help to ensure that system parameters remain identifiable. Although here we advance model (1) mainly as a tool for approximating unknown dynamical systems, it may be interpreted as a neural rate model [e.g. 37, 38], with $\mathbf{A}$ the units' passive time constants, $\mathbf{W}$ the synaptic coupling matrix, and $\varphi(\mathbf{z})$ a current/voltage to spike rate transfer function which for cortical pyramidal cells is often non-saturating and close to a ReLU within the physiologically relevant regime [e.g. 39].

The observed time series are generated from the ReLU-transformed latent states (eq. 1) through a linear-Gaussian model:

$$(2) \quad \mathbf{x}_t = \mathbf{B}\varphi(\mathbf{z}_t) + \boldsymbol{\eta}_t, \quad \boldsymbol{\eta}_t \sim N(\mathbf{0}, \boldsymbol{\Gamma})$$

where $\mathbf{x}_t$ are the observed *N*-dimensional measurements at time *t* generated from $\mathbf{z}_t$, $\mathbf{B} \in \mathbb{R}^{N \times M}$ is a matrix of regression weights (factor loadings), and $\mathbf{\eta}_t$ denotes a Gaussian white observation noise process with diagonal covariance matrix $\mathbf{\Gamma}$.

Thus, the model is specified by the set of parameters $\mathbf{\theta} = \{\mathbf{\mu}_0, \mathbf{A}, \mathbf{W}, \mathbf{C}, \mathbf{h}, \mathbf{B}, \mathbf{\Gamma}, \mathbf{\Sigma}\}$, and we are interested in recovering $\mathbf{\theta}$ as well as the posterior distribution $p(\mathbf{Z}|\mathbf{X})$ over the latent state path $\mathbf{Z} = \{\mathbf{z}_{1:T}\}$ from the experimentally observed time series $\mathbf{X} = \{\mathbf{x}_{1:T}\}$ and experimental inputs $\mathbf{S} = \{\mathbf{s}_{1:T}\}$. In the following, we will sometimes use the notation $\mathbf{\theta}_{lat} = \{\mathbf{\mu}_0, \mathbf{A}, \mathbf{W}, \mathbf{C}, \mathbf{h}, \mathbf{\Sigma}\}$ and $\mathbf{\theta}_{obs} = \{\mathbf{B}, \mathbf{\Gamma}\}$ to exclusively refer to parameters in the evolution or observation equation, respectively.

**Observation model for BOLD time series**

An appealing feature of the SSM framework is that different measurement modalities and properties can be accommodated by connecting different observation models to the same latent model. In order to apply our model to fMRI time series, we need only to adapt observation eq. 2 to meet the distributional assumptions and temporal filtering of the blood-oxygen-level dependent (BOLD) signal, while retaining process eq. 1 with its universal approximation capabilities. In contrast to electrophysiological measurements, BOLD time-series are a strongly filtered, highly smoothed version of some underlying neural process, only accessible through the hemodynamic response function (HRF) [e.g. 40]. Hence, we modified the observation model (eq. 2) such that the observed BOLD signal is generated from the latent states (eq. 1) through a linear-Gaussian model with HRF convolution:

$$(3) \quad \mathbf{x}_t = \mathbf{B}(hrf * \mathbf{z}_{\tau:t}) + \mathbf{J}\mathbf{r}_t + \mathbf{\eta}_t, \quad \mathbf{\eta}_t \sim N(\mathbf{0}, \mathbf{\Gamma})$$

where $\mathbf{x}_t$ are the observed BOLD responses in *N* voxels at time *t* generated from $\mathbf{z}_{\tau:t}$ (concatenated into one vector and convolved with the hemodynamic response function). We also added nuisance predictors $\mathbf{r}_t \in \mathbb{R}^P$, which account for artifacts caused, e.g., by movements. $\mathbf{J} \in \mathbb{R}^{N \times P}$ is the coefficient matrix of these nuisance variables, and $\mathbf{B}$, $\mathbf{\Gamma}$ and $\mathbf{\eta}_t$ are the same as in eq. 2. Hence, the observation model takes the typical form of a General Linear Model for BOLD signal analysis as, e.g., implemented in the statistical parametric mapping (SPM) framework [40]. Note that while nuisance variables are assumed to directly blur into the observed signals (they do not affect the neural dynamics but rather the recording process), external stimuli presented to the subjects are, in contrast, assumed to exert their effects through the underlying neuronal dynamics (eq. 1). Thus, the fMRI PLRNN-SSM (termed 'PLRNN-BOLD-SSM') is now specified by the set of parameters $\mathbf{\theta} = \{\mathbf{\mu}_0, \mathbf{A}, \mathbf{W}, \mathbf{C}, \mathbf{h}, \mathbf{B}, \mathbf{J}, \mathbf{\Gamma}, \mathbf{\Sigma}\}$. Model inference is performed through a type of Expectation-Maximization (EM) algorithm (see Methods and full derivations in supporting file S1 Text).

One complication here is that the observations in eq. 3 do not just depend on the current state $\mathbf{z}_t$ as in a conventional SSM, but on a set of states $\mathbf{z}_{\tau:t}$ across several previous time steps. This severely complicates standard solution techniques for the E-step like extended or unscented Kalman filtering [41]. Our E-step procedure [cf. 31], however, combines a global Laplace approximation with an efficient iterative (fixed point-type) mode search algorithm that exploits the sparse, block-banded structure of the involved covariance (inverse Hessian) matrices, which is more easily adapted for the current situation with longer-term temporal dependencies (see Methods sect. 'Model specification and inference' & S1 Text for further details).

**Stepwise initialization and training protocol**

The EM-algorithm aims to compute (in the linear case) or approximate the posterior distribution $p(\mathbf{Z}|\mathbf{X})$ of the latent states given the observations in the E-step, in order to maximize the expected joint log-likelihood $\mathrm{E}_{q(\mathbf{Z}|\mathbf{X})}[\log p_{\boldsymbol{\theta}}(\mathbf{Z}, \mathbf{X})]$ with respect to the unknown model parameters $\boldsymbol{\theta}$ under this approximate posterior $q(\mathbf{Z}|\mathbf{X}) \approx p(\mathbf{Z}|\mathbf{X})$ in the M-step (by doing so, a lower bound of the log-likelihood $\log p(\mathbf{X}|\boldsymbol{\theta}) \geq \mathrm{E}_q[\log p(\mathbf{Z}, \mathbf{X})] - \mathrm{E}_q[\log q(\mathbf{Z}|\mathbf{X})]$ is maximized, see Methods sect. 'Parameter estimation' & S1 Text). This does not by itself guarantee that the latent system on its own, as represented by the prior distribution $p_{\boldsymbol{\theta}_{lat}}(\mathbf{Z})$, provides a good incarnation of the true but unobserved DS that generated the observations $\mathbf{X}$. As for any nonlinear neural network model, the log-likelihood landscape for our model is complicated and usually contains many local modes, very flat and saddle regions [42-45]. Since $\mathrm{E}_q[\log p(\mathbf{Z}, \mathbf{X})] = \mathrm{E}_q[\log p(\mathbf{X}|\mathbf{Z})] + \mathrm{E}_q[\log p(\mathbf{Z})]$, with the expectation taken across $q(\mathbf{Z}|\mathbf{X}) \approx p(\mathbf{Z}|\mathbf{X}) \propto p(\mathbf{X}|\mathbf{Z})p(\mathbf{Z})$, the inference procedure may easily get stuck in local maxima in which high likelihood values are attained by finding parameter and state configurations which overemphasize fitting the observations, $p(\mathbf{X}|\mathbf{Z})$, rather than capturing the underlying dynamics in $p(\mathbf{Z})$ (eq. 1; see Methods for more details). To address this issue, we here propose a step-wise training by annealing protocol (termed 'PLRNN-SSM-anneal', Algorithm-1 in Methods) which systematically varies the trade-off between fitting the observations (maximizing $p(\mathbf{X}|\mathbf{Z})$; eqns. 2-3) as compared to fitting the dynamics ($p(\mathbf{Z})$; eq. 1) in successive optimization steps [see also 46]. In brief, while early steps of the training scheme prioritize the fit to the observed measurements through the observation (or 'decoder') model $p(\mathbf{X}|\mathbf{Z})$ (eqns. 2-3), subsequent annealing steps shift the burden of reproducing the observations onto the latent model $p(\mathbf{Z})$ (eq. 1) by, at some point, fixing the observation parameters $\boldsymbol{\theta}_{obs}$, and then enforcing the temporal consistency within the latent model equations (as demanded by eq. 1) by gradually boosting the contribution of this term to the log-likelihood (see Methods).

**Evaluation of training protocol**

We examined the performance of this annealing protocol in terms of how well the inferred model was capable of recovering the true underlying dynamics of the Lorenz system. This 3-dimensional benchmark system (equations and parameter values used given in Fig 4 legend), conceived by Edward Lorenz in 1963 to describe atmospheric convection [47], exhibits chaotic behavior in certain regimes (see, e.g., Fig 4A). We measured the quality of DS reconstruction by the Kullback-Leibler divergence $KL_{\mathbf{x}}\big(p_{true}(\mathbf{x}), p_{gen}(\mathbf{x}|\mathbf{z})\big)$ between the spatial probability distributions $p_{true}(\mathbf{x})$ over observed system states in $\mathbf{x}$-space from trajectories produced by the (true) Lorenz system and $p_{gen}(\mathbf{x}|\mathbf{z})$ from trajectories generated by the trained PLRNN-SSM ($KL_{\mathbf{x}}$, in the following refers to this divergence evaluated in observation space, see eq. 9 in Methods, where $\widetilde{KL}_{\mathbf{x}}$ denotes a normalized version of this measure; see Fig 1 and Methods sect. 'Reconstruction of benchmark dynamical systems' for details). Hence, importantly, our measure compares the dynamical behavior in *state space*, i.e. focuses on the agreement between attractor (or, more generally trajectory) *geometries*, similar in spirit to the delay embedding theorems (which ensure topological equivalence) [49-51], instead of comparing the fit directly on the time series themselves which can be highly misleading for chaotic systems because of the exponential divergence of nearby trajectories [e.g. 48], as illustrated in Fig 2A. Note that for a (deterministic, autonomous) dynamical system the flow at each point in state space is uniquely determined [e.g. 24] and induces a specific spatial distribution of states, in this sense translates the temporal dynamics into a specific spatial geometry. Fig 2B gives examples where our measure $\widetilde{KL}_\mathbf{x}$ correctly indicates whether the Lorenz attractor geometry (and hence the underlying dynamical system) was properly mapped by a trained PLRNN, while a direct evaluation of the time series fit (incorrectly) indicated the contrary.

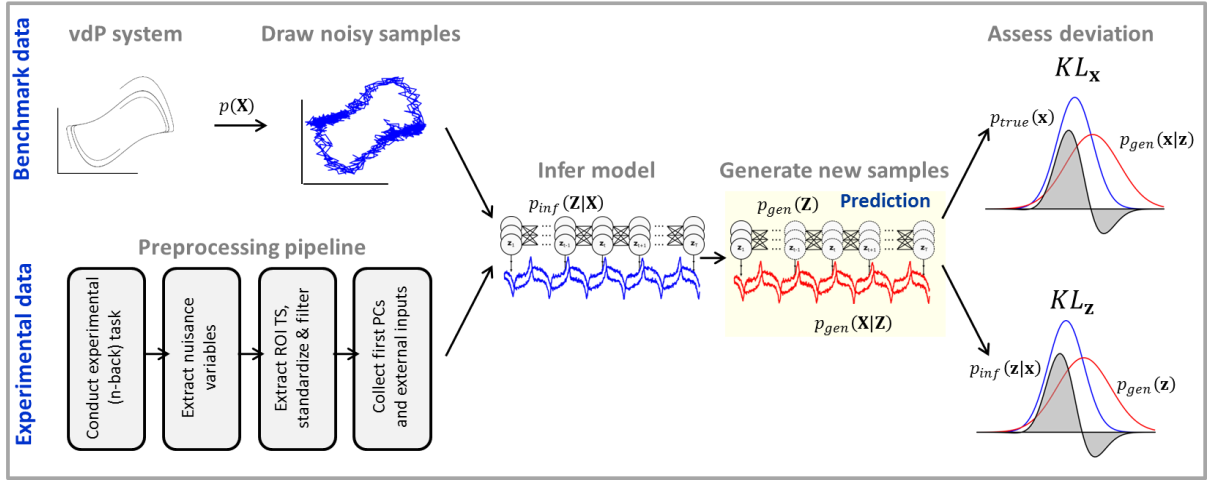

**Fig 1. Analysis pipeline.** Top: Analysis pipeline for simulated data. From the two benchmark systems (van der Pol and Lorenz systems), noisy trajectories were drawn and handed over to the PLRNN-SSM inference algorithm. With the inferred model parameters, completely new trajectories were generated and compared to the state space distribution over true trajectories via the Kullback-Leibler divergence $KL_\mathbf{x}$ (see eq. 9). Bottom: analysis pipeline for experimental data. We used preprocessed fMRI data from human subjects undergoing a classic working memory n-back paradigm. First, nuisance variables, in this case related to movement, were collected. Then, time series obtained from regions of interest (ROI) were extracted, standardized, and filtered (in agreement with the study design). From these preprocessed time series, we derived the first principle components and handed them to the inference algorithm (once including and once excluding variables indicating external stimulus presentations during the experiment). With the inferred parameters, the system was then run freely to produce new trajectories which were compared to the state space distribution from the inferred trajectories via the Kullback-Leibler divergence $KL_\mathbf{z}$ (see eq. 11).

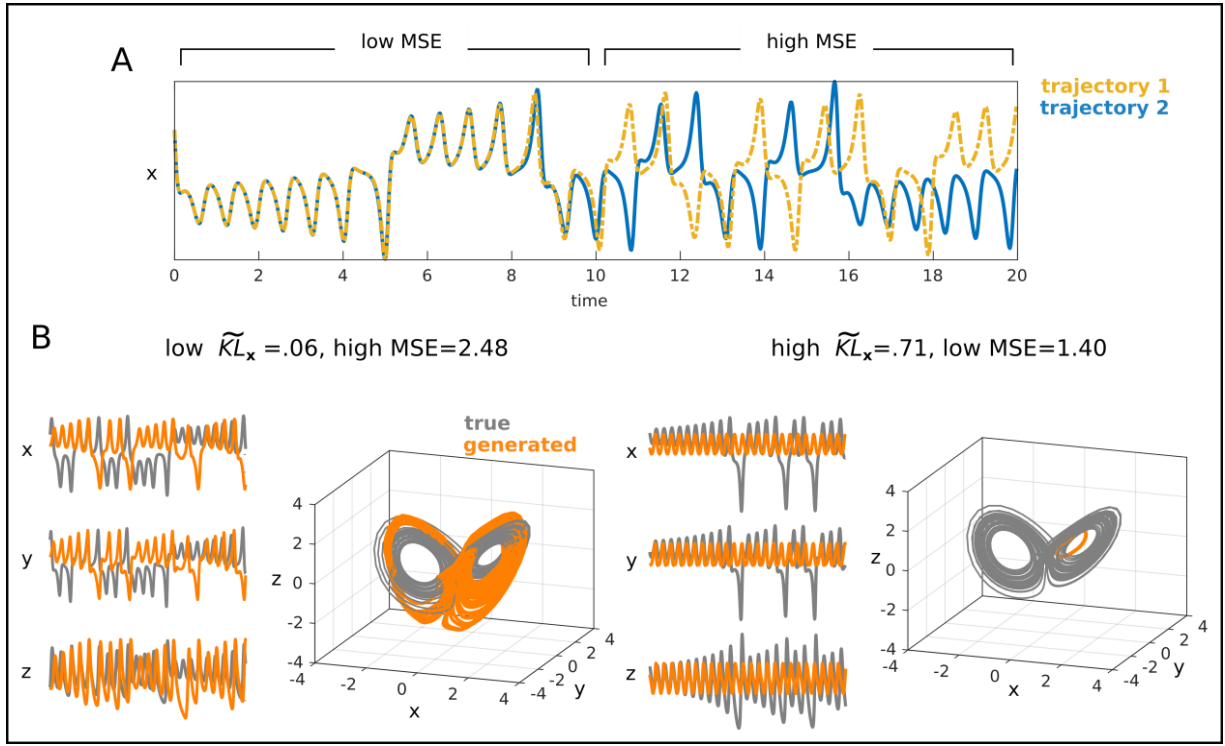

**Fig 2. Illustration of DS reconstruction measures defined in state space ($\widetilde{KL}_\mathbf{x}$) vs. on the time series (mean squared error; MSE)**. A. Two noise-free time series from the Lorenz equations started from slightly different initial conditions. Although initially the two time series (blue and yellow) stay closely together (low MSE), they then quickly diverge yielding a very large discrepancy in terms of the MSE, although truly they come from the very same system with the very same parameters. These problems will be aggravated once noise is added to the system and initial conditions are not tightly matched (as almost impossible for systems observed empirically), rendering any measure based on direct matching between time series a relatively poor choice for assessing dynamical systems reconstruction except for a couple of initial time steps. B. Example time series and state spaces from trained PLRNN-SSMs which capture the chaotic structure of the Lorenz attractor quite well (left) or produce rather a simple limit cycle but not chaos (right). The dynamical reconstruction quality is correctly indicated by $\widetilde{KL}_\mathbf{x}$ (low on the left but high on the right), while the MSE between true (orange) and generated (gray) time series, on the contrary, would wrongly suggest that the right reconstruction (MSE = 1.4) is better than the one on the left (MSE = 2.48).

For evaluating our specific training protocol (termed 'PLRNN-SSM-anneal', Algorithm-1 in Methods), trajectories of length $T=1000$ were drawn with process noise ($\sigma^2=.3$) from the Lorenz system and handed to the inference algorithm with $M=\{8, 10, 12, 14\}$ latent states (for statistics, a total of 100 such trajectories were simulated and model fits carried out on each). Models were trained through 'PLRNN-SSM-anneal' and compared to models trained from random initial conditions (termed 'PLRNN-SSM-random') in which parameters were randomly initialized (see Fig 3).

In general, the PLRNN-SSM-anneal protocol significantly decreased the normalized KL divergence $\widetilde{KL}_\mathbf{x}$ (eq. 9) and increased the joint log-likelihood when compared to the PLRNN-SSM-random initialization scheme (see Fig 3A,B, independent $t$-test on $\widetilde{KL}_\mathbf{x}$: $t(686)=-16.3$, $p<.001$, and on the expected joint log-likelihood: $t(640)=11.32$, $p<.001$). More importantly though, the PLRNN-SSM-anneal protocol produced more estimates for which $\widetilde{KL}_\mathbf{x}$ was in a regime in which the chaotic attractor could be well reconstructed (see Fig 4, grey shaded area indicates $\widetilde{KL}_\mathbf{x}$ values for which the chaotic attractor was reproduced). Furthermore, the expected joint log-likelihood increased (Fig 3D) while $KL_\mathbf{x}$ decreased (Fig 3C) over the distinct training steps of the PLRNN-SSM-anneal protocol, indicating that each step further enhances the solution quality. $\widetilde{KL}_\mathbf{x}$ and the normalized log-likelihood were, however, only moderately correlated ($r=-27$, $p<.001$), as expected based on the formal considerations above (sect. 'Stepwise initialization and training protocol').

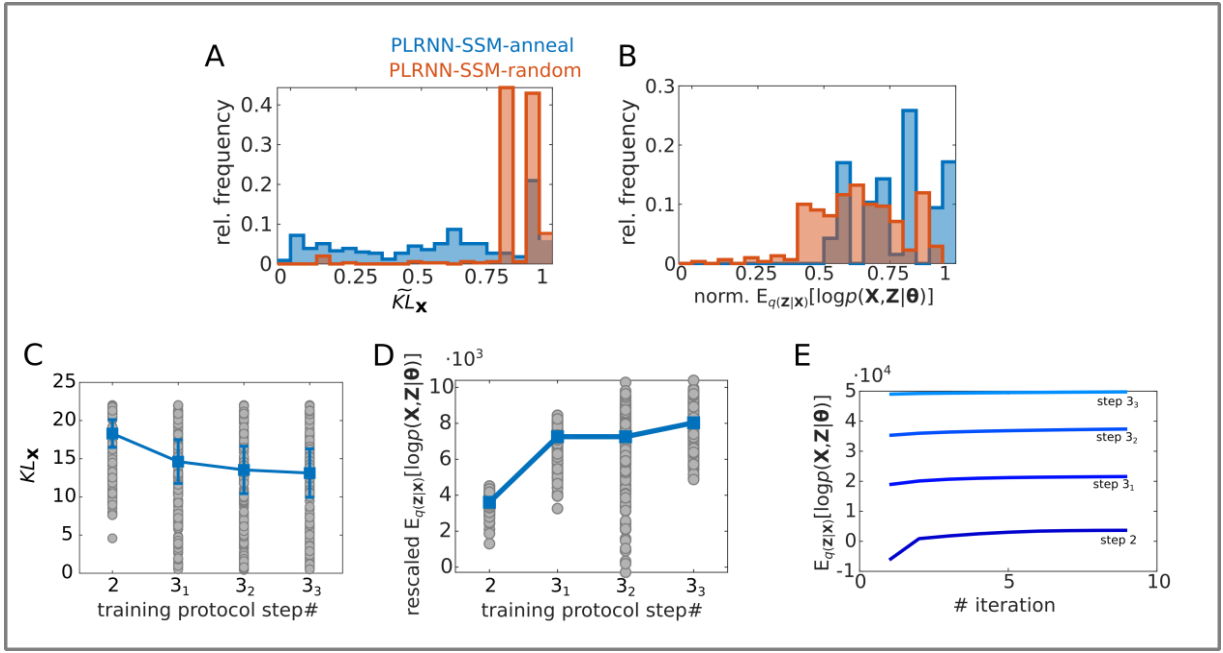

**Fig 3. Evaluation of stepwise training protocol on chaotic Lorenz attractor.** A. Relative frequency of normalized KL divergences evaluated on the observation space ($\widetilde{KL}_\mathbf{x}$) after running the EM algorithm with the PLRNN-SSM-anneal (blue) and PLRNN-SSM-random (red) protocols on 100 distinct trajectories drawn from the Lorenz system (with $T=1000$, and $M=8, 10, 12, 14$). B. Same as A for normalized expected joint log-likelihood $\mathbb{E}_{q(\mathbf{Z}|\mathbf{X})}[\log p(\mathbf{X},\mathbf{Z}|\boldsymbol{\theta})]$ (see S1 Text eq. 1). C. Decrease in $KL_\mathbf{x}$ over the distinct training steps of 'PLRNN-SSM-anneal' (see Algorithm-1; the first step refers to a LDS initialization and was removed). D. Increase in (rescaled) expected joint log-likelihood across training steps 2-$3_{1-3}$ in 'PLRNN-SSM-anneal'. Since the protocol partly works by systematically scaling down $\boldsymbol{\Sigma}$, for comparability the log-likelihood after each step was recomputed (rescaled) by setting $\boldsymbol{\Sigma}$ to the identity matrix. E. Representative example of joint log-likelihood increase during the EM iterations of the individual training steps 2-$3_{1-3}$ for a single Lorenz trajectory. Unstable system estimates and likelihood values<-$10^3$ were removed from all figures for visualization purposes.

**Reconstruction of benchmark dynamical systems**

After establishing an efficient training procedure designed to enforce recovery of the underlying DS by the prior model (eq. 1), we more formally evaluated dynamical reconstructions on the chaotic Lorenz system and on the van der Pol (vdP) nonlinear oscillator. The vdP oscillator with nonlinear dampening is a simple 2-dimensional model for electrical circuits consisting of vacuum tubes [52] (equations given in Fig 4). Fig 4 illustrates its flow field in the plane, together with several trajectories converging to the system's limit cycle (note that training was always performed on samples of the time series, not on the generally unknown flow field!).

As for the Lorenz system, we drew 100 time series samples of length $T=1000$ with process noise ($\sigma^2=.1$) using Runge-Kutta numerical integration, and handed each of those over to a separate PLRNN-SSM inference run with $M=\{8, 10, 12, 14\}$ latent states. As above, reconstruction performance was assessed in terms of the (normalized) KL divergence $\widetilde{KL}_\mathbf{x}$ (eq. 9) between the distributions over true and generated states in state space. In addition, for the chaotic attractor, the absolute difference between Lyapunov exponents [e.g. 51] from the true vs. the PLRNN-SSM-generated trajectories was assessed, as another measure of how well hallmark dynamical characteristics of the chaotic Lorenz system had been captured. For the vdP (non-chaotic) oscillator, we instead assessed the correlation between the power spectrum of the true and the generated trajectories (see Methods sect. 'Reconstruction of benchmark dynamical systems').

Overall, our PLRNN-SSM-anneal algorithm managed to recover the nonlinear dynamics of these two benchmark systems (see Fig 4). The inferred PLRNN-SSM equations reproduced the 'butterfly' structure of the somewhat challenging chaotic attractor very well (Fig 4D). The $\widetilde{KL}_\mathbf{x}$ measure effectively captured this reconstruction quality, with PLRNN reconstructions achieving values below $\widetilde{KL}_\mathbf{x} \approx .4$ agreeing well with the Lorenz attractor's 'butterfly' structure as assessed by visual inspection (see Fig 4B). At the same time, for this range of $\widetilde{KL}_\mathbf{x}$ values the deviation between Lyapunov exponents of the true and generated Lorenz system was generally very low (see Fig 4C, grey shaded area). If we accept this value as an indicator for successful reconstruction, our algorithm was successful in 15%, 24%, 35%, and 28% of all samples for *M*=8, 10, 12, and 14 states, respectively (note that our algorithm had access only to rather short time series of *T=1000*, to create a situation comparable to the fMRI data studied later). When examining the dependence of $\widetilde{KL}_x$ on the number of latent states across a larger range in more detail, $M \approx 16$ turned out to be optimal for this setting (S1 Fig). Importantly and in contrast to most previous studies, note we requested full independent generation of the original attractor object from the once trained PLRNN. That is, we neither 'just' evaluated the posterior $p(\mathbf{Z}|\mathbf{X})$ conditioned on the actual observations (as e.g. in [53], or [36]), nor did we 'just' assess predictions a couple of time steps ahead (as, e.g., in [31]), but rather defined a much more ambitious goal for our algorithm.

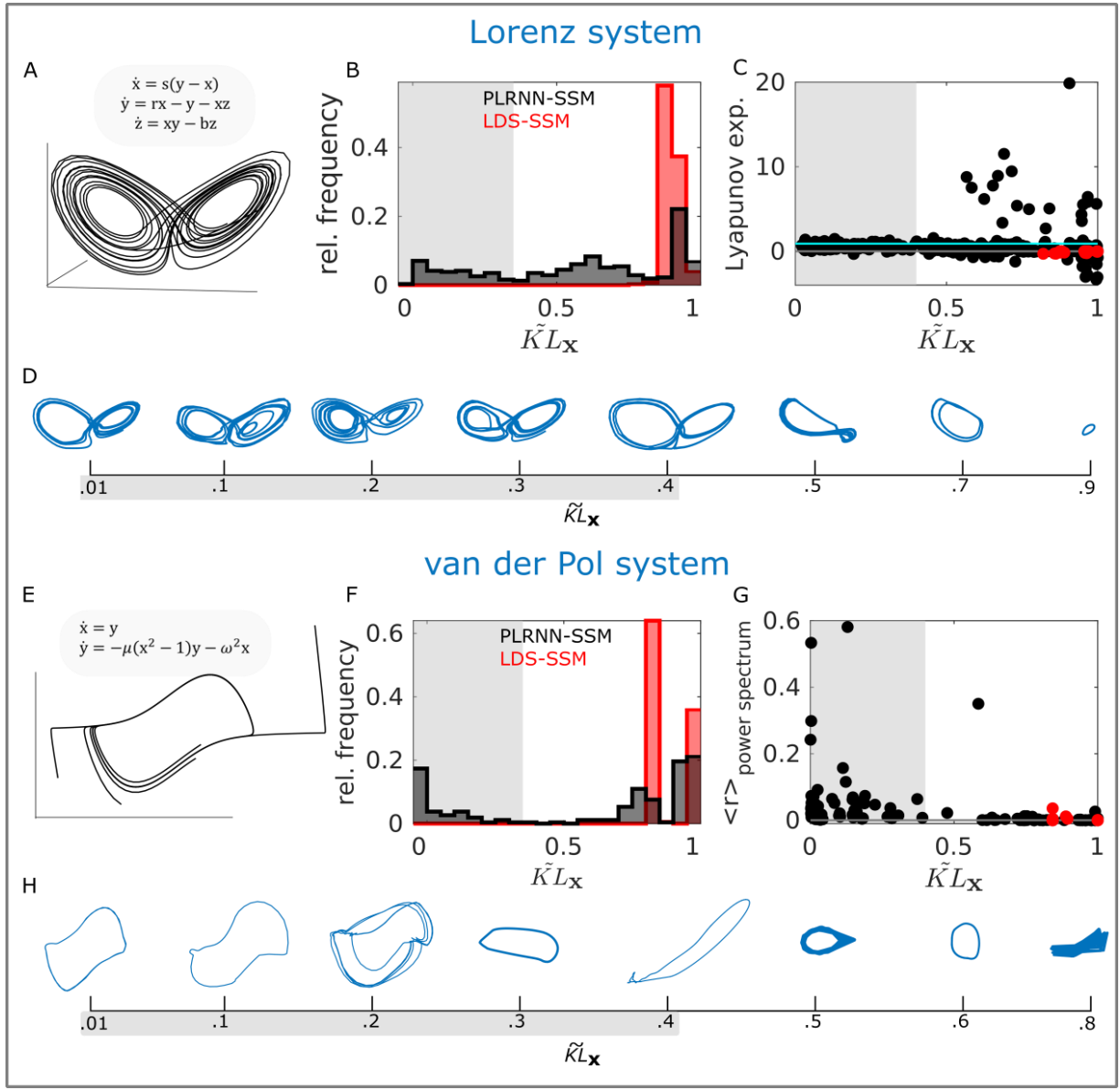

**Fig 4. Evaluation of training protocol and KL measure on dynamical systems benchmarks.** A. True trajectory from chaotic Lorenz attractor (with parameters s=10, r=28, b=8/3). B. Distribution of $\widetilde{KL}_\mathbf{x}$ (eq. 9) across all samples, binned at .05, for PLRNN-SSM (black) and LDS-SSM (red). For the PLRNN-SSM, around 26% of these samples (grey shaded area, pooled across different numbers of latent states $M$) captured the butterfly structure of the Lorenz attractor well (see also D). Unsurprisingly, the LDS completely failed to reconstruct the Lorenz attractor. C. Estimated Lyapunov exponents for reconstructed Lorenz systems for PLRNN-SSM (black) and LDS-SSM (red) (estimated exponent for true Lorenz system ≈.9, cyan line). A significant positive correlation between the absolute deviation in Lyapunov exponents for true and reconstructed systems with $\widetilde{KL}_\mathbf{x}$ ($r=.27$, $p<.001$) further supports that the latter measures salient aspects of the nonlinear dynamics in the PLRNN-SSM (for the LDS-SSM, all of these empirically determined Lyapunov exponents were either < 0, as indicative of convergence to a fixed point, or at least very close to 0, light-gray line). D. Samples of PLRNN-generated trajectories for different $\widetilde{KL}_\mathbf{x}$ values. The grey shaded area indicates successful estimates. E. True van der Pol system trajectories (with μ=2 and ω=1). F. Same as in B but for van der Pol system. G. Correlation of the spectral density between true and reconstructed van der Pol systems for the PLRNN-SSM (black) and LDS-SSM (red). A significant negative correlation for the PLRNN-SSM between the agreement in the power spectrum (high values on y-axis) and $\widetilde{KL}_\mathbf{x}$ again supports that the normalized KL divergence defined across state space (eq. 9) captures the dynamics (we note that measuring the correlation between power spectra comes with its own problems, however). For the LDS-SSM, in contrast, all power-spectra correlations and $\widetilde{KL}_\mathbf{x}$ measures were poor. H. Same as in D for van der Pol system. Note that even reconstructed systems with high $\widetilde{KL}_\mathbf{x}$ values may capture the limit cycle behavior and thus the basic topological structure of the underlying true system (in general, the 2-

dimensional vdP system is likely easier to reconstruct than the chaotic Lorenz system; vice versa, low $\widetilde{KL}_\mathbf{x}$ values do not generally ascertain that the reconstructed system exhibits the same frequencies).

For the vdP system, our inference procedure yielded agreeable results in 20%, 31%, 25%, and 35% of all samples for $M$=8, 10, 12, and 14 states, respectively (grey shaded area in Fig 4F), with $M$=14 about optimal for this setting (S1 Fig). Furthermore, around 50% of all estimates generated stable limit cycles and hence a topologically equivalent attractor object in state space, although these limit cycles varied a lot in frequency and amplitude compared to the true oscillator. Like for the Lorenz system, the $\widetilde{KL}_\mathbf{x}$ measure generally served as a good indicator of reconstruction quality (see Fig 4H), particularly when combined with the power spectrum correlation (Fig 4G), although low $\widetilde{KL}_\mathbf{x}$ values did not always guarantee and high values did not exclude the retrieval of a stable limit cycle.

As noted in the Introduction, a linear dynamical system (LDS) is inherently (mathematically) incapable of producing more complex dynamical phenomena like limit cycles or chaos. To explicitly illustrate this, we ran the same training procedure (Algorithm-1) on a *linear* state space model (LDS-SSM) which we created by simply swapping the ReLU nonlinearity $\varphi(\mathbf{z}) = \max(\mathbf{z}, 0)$ with the linear function $\varphi(\mathbf{z}) = \mathbf{z}$ in eqns. 1-2. As expected, this had a dramatic effect on the system's capability to capture the true underlying dynamics, with $\widetilde{KL}_\mathbf{x}$ close to 1 in most cases for both the Lorenz (Fig 4B,C) and the vdP (Fig 4F,G) equations. Even for the simpler (but nonlinear) oscillatory vdP system, LDS-SSM would at most produce damped (and linear, harmonic) oscillations which decay to a fixed point over time (Fig 5A).

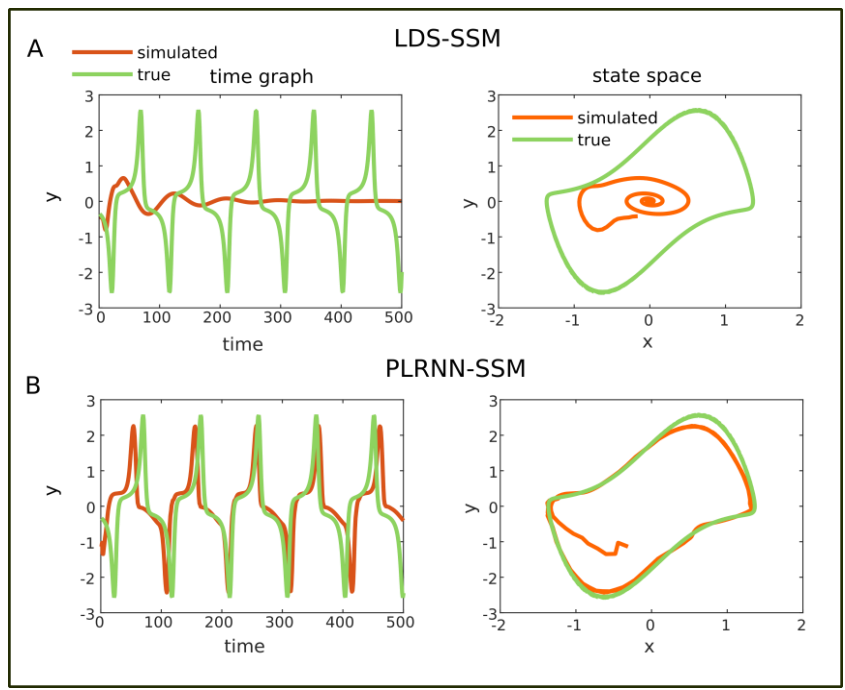

**Fig 5. Example time series from an LDS-SSM and a PLRNN-SSM trained on the vdP system**. A. Example time graph (left) and state space (right) for a trajectory generated by an LDS-SSM (solid lines) trained on the vdP system (true vdP trajectories as dashed lines). Trajectories from a LDS will almost inevitably decay toward a fixed point over time (or diverge). B. Trajectories generated by a trained PLRNN-SSM, in contrast, closely follow the vdP-system's original limit cycle.

**Reconstruction of experimental data**

We next tested our PLRNN inference scheme, with a modified observation model that takes the hemodynamic response filtering into account (PLRNN-BOLD-SSM; see sect. 'Observation model for BOLD time series'), on a previously published experimental fMRI data set [54]. In brief, the experimental paradigm assessed three cognitive tasks presented within repeated blocks, two variants of the well-established working memory (WM) n-back task: a 1-back continuous delayed response task (CDRT), a 1-back continuous matching task (CMT), and a (0-back control) choice reaction task (CRT). Exact details on the experimental paradigm, fMRI data acquisition, preprocessing, and sample information can be found in [54]. From these data obtained from 26 subjects, we preselected as time series the first principle component from each of 10 bilateral regions identified as relevant to the n-back task in a previous meta-analysis [55]. These time series along with the individual movement vectors obtained from the SPM realignment procedure (see also Methods sect. 'Data acquisition and preprocessing') were given to the inference algorithm for each subject: Models with $M=\{1,...,10\}$ latent states were inferred twice: once explicitly including, and once excluding external (experimental) inputs (i.e., in the latter analysis, the model had to account for fluctuations in the BOLD signal all by itself, without information about changes in the environment).

For experimentally observed time series, unlike for the benchmark systems, we do not know the ground truth (i.e., the true data generating process), and generally do not have access to the complete true state space either (but only to some possibly incomplete, nonlinear projection of it). Thus, we cannot determine the agreement between generated and true distributions directly in the space of observables, as we could for the benchmark systems. Therefore we use a proxy: If the prior dynamics is close to the true system which generated the experimental observations, and those represent the true dynamics well (at the very least, they are the best information we have), then the distribution of latent states constrained by the data, i.e. $p(\mathbf{Z}|\mathbf{X})$, should be a good representative of the distribution over latent states generated by the prior model on its own, i.e. $p(\mathbf{Z})$. Hence, our proxy for the reconstruction quality is the KL divergence $KL_{\mathbf{z}}\left(p_{inf}(\mathbf{z}|\mathbf{x}), p_{gen}(\mathbf{z})\right)$ ($KL_{\mathbf{z}}$ for short, or, when normalized, $\widetilde{KL}_{\mathbf{z}}$; see eq. 11 in Methods) between the posterior (inferred) distribution $p_{inf}(\mathbf{z}|\mathbf{x})$ over latent states $\mathbf{z}$ conditioned on the experimental data $\mathbf{x}$, and the spatial distribution $p_{gen}(\mathbf{z})$ over latent states as generated by the model's prior (governing the free-running model dynamics; we use capital letters, $\mathbf{Z}$, and lowercase letters, $\mathbf{z}$, to distinguish between full trajectories and single vector points in state space, respectively). Note that the latent space defines a complete state space as we have that complete model available (also note that our measure, as before, assesses the agreement in *state space*, not the agreement between time series).

For the benchmark systems, our proposed proxy $KL_{\mathbf{z}}$ was well correlated with the KL divergence $KL_{\mathbf{x}}$ assessed directly in the complete observation space, i.e., between true and generated distributions (Fig 6A, $r=.72$ on a logarithmic scale, $p<.001$; likewise, $KL_{\mathbf{z}}\left(p_{inf}(\mathbf{z}|\mathbf{x}), p_{gen}(\mathbf{z})\right)$ and $KL_{\mathbf{z}}\left(p_{gen}(\mathbf{z}), p_{inf}(\mathbf{z}|\mathbf{x})\right)$ were generally correlated highly; $r>.9$, $p<.001$). Moreover, although especially for chaotic systems we would not necessarily expect a good fit between observed or inferred and generated time series [c.f. 48], $\widetilde{KL}_\mathbf{z}$ on the latent space turned out to be significantly related to the correlation between inferred and generated latent state *series* in our case (on a logarithmic scale, see Fig 6B). That is, lower $\widetilde{KL}_\mathbf{z}$ values were associated with a better match of inferred and generated state trajectories.

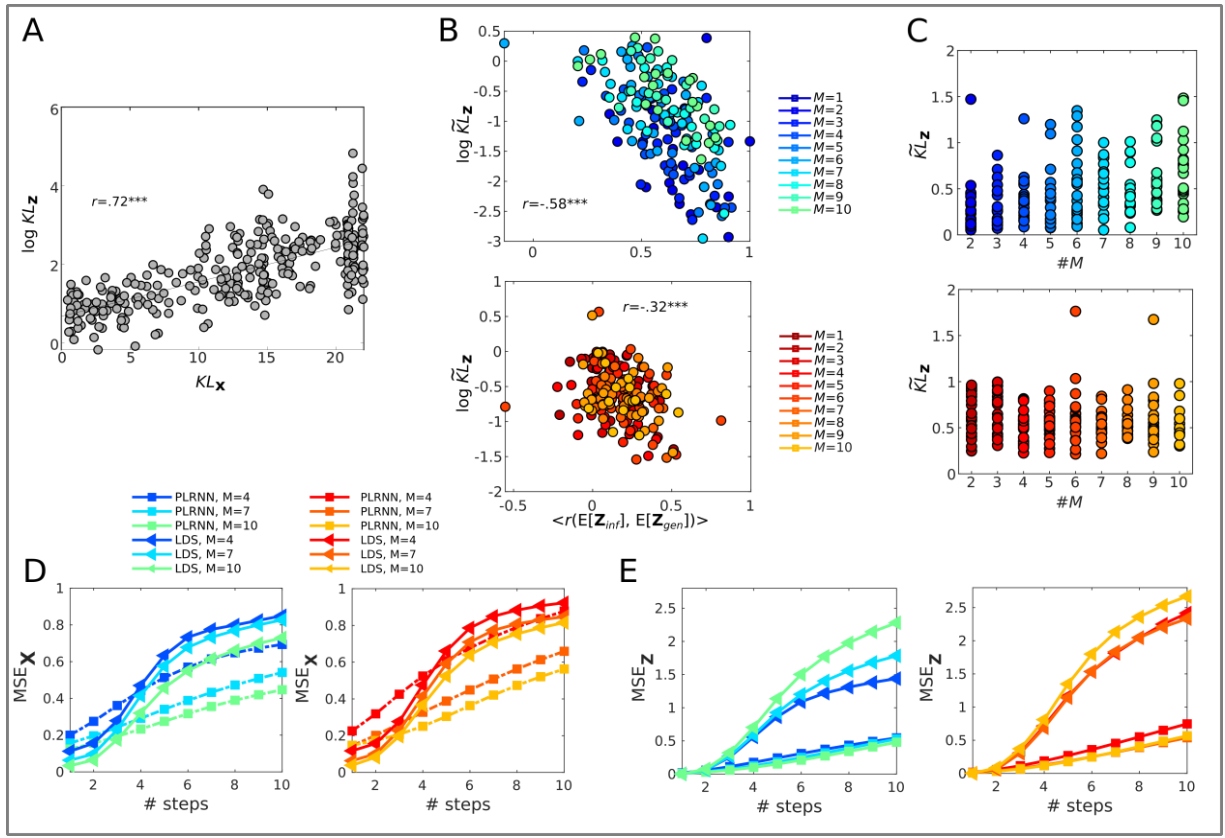

**Fig 6. Model evaluation on experimental data.** A. Association between KL divergence measures on observation ($KL_\mathbf{x}$) vs. latent space ($KL_\mathbf{z}$) for the Lorenz system; y-axis displayed in log-scale. B. Association between $\widetilde{KL}_\mathbf{z}$ (eq. 11; in log scale) and correlation between generated and inferred state series for models with inputs (top, displayed in shades of blue for $M$=2...10), and models without inputs (bottom, displayed in shades of red for $M$=2...10). C. Distributions of $\widetilde{KL}_\mathbf{z}$ (y-axis) in an experimental sample of n=26 subjects for different latent state dimensions (x-axis), for models including (top) or excluding (bottom) external inputs. D. Mean squared error (MSE) between generated and true observations for the PLRNN-BOLD-SSM (dashed-squares) and the LDS-BOLD-SSM (solid-triangles) as a function of ahead-prediction step for models including (left) or excluding (right) external inputs. The PLRNN-BOLD-SSM starts to robustly outperform the LDS-BOLD-SSM for predictions of observations more than about 3 time steps ahead, the latter in contrast to the former exhibiting a strongly nonlinear rise in prediction errors from that time step onward. The LDS-BOLD-SSM also does not seem to profit as much from increasing the latent state dimensionality. E. Same as D for the MSE between generated and inferred states as a function of ahead-prediction step, showing that the comparatively sharp rise in prediction errors for the LDS-BOLD-SSM in contrast to the PLRNN-BOLD-SSM is accompanied by a sharp increase in the discrepancy between generated and inferred state trajectories after the 3rd prediction step. Unstable system estimates were removed from D and E.

This tight relation was particularly pronounced in models including external inputs (Fig 6B blue, top). This is expected, as in this case the internal dynamics are reset or partly driven by the external inputs, which will therefore induce correlations between directly inferred and freely generated trajectories. Thus, overall, $KL_\mathbf{z}$ was slightly lower for models including external inputs as compared to autonomous models (see also Fig 6C). One simple but important conclusion from this is that knowledge about additional external inputs and the experimental task structure may (strongly) help to recover the true underlying DS. This was also evident in the mean squared error on *n*-step ahead projections of generated as compared to true data (Fig 6D), i.e. when comparing predicted observations from the PLRNN-BOLD-SSM run freely for *n* time steps to the true observations (once again we stress, however, that a measure evaluated directly on the time series may not necessarily give a good intuition about whether the underlying DS has been captured well; see also Fig 2). Accuracy of *n*-step-ahead predictions also generally improved with increasing number of latent state dimensions, that is, adding latent states to the model appeared to enhance the dynamical reconstruction within the range studied here.

In contrast to the PLRNN-BOLD-SSM, the performance of the LDS-SSM with the same BOLD observation model (termed LDS-BOLD-SSM), and trained according to the same protocol (Algorithm-1, see also previous section), quickly decayed after about only three prediction time steps (Fig 6D), clearly below the prediction accuracy achieved by the PLRNN-BOLD-SSM for which the decay was much more linear. Interestingly, this comparatively sharp drop in prediction accuracy for the LDS-BOLD-SSM, unlike the PLRNN-BOLD-SSM, was accompanied by a similarly sharp rise in the discrepancy between generated and inferred latent state trajectories (Fig 6E), which was not apparent for the PLRNN-BOLD-SSM. This suggests that the rise in LDS-BOLD-SSM prediction errors is directly related to the model's inability to capture the underlying system in its *generative* dynamics (while the inferred latent states may still provide reasonable fits), and – moreover – that the agreement between inferred and generated latent states is indeed a good indicator of how well this goal of reconstructing dynamics has been achieved. The linear model's failure to capture the underlying dynamics was also evident from the fact that its generated trajectories often quickly converged to fixed points (Fig 7C), while the trained PLRNNs often mimicked the oscillatory activity found in the real data in their generative behavior (Fig 7B).

Moreover, we observed that a PLRNN-BOLD model fit directly to the observations (as one would, e.g., do for an ARMA model; see Methods), i.e. essentially lacking latent states, was much worse in forecasting the time series than either the PLRNN-BOLD-SSM or the LDS-BOLD-SSM, with predictions errors on average above 3.28 even for just a single time step ahead, either when external inputs were absent (MSE > 2.79 for 1-step) or present (MSE > 3.77 for 1-step), as compared to the results for the latent variable models in Fig. 6D. On top, they produced a large number of unstable solutions (35% and 46%, respectively). This suggests that the latent state structure is absolutely necessary for reconstructing the dynamics, perhaps not surprisingly so given that the whole motivation behind delay embedding techniques in nonlinear dynamics is that the true attractor geometries are almost never accessible directly in the observation space [51].

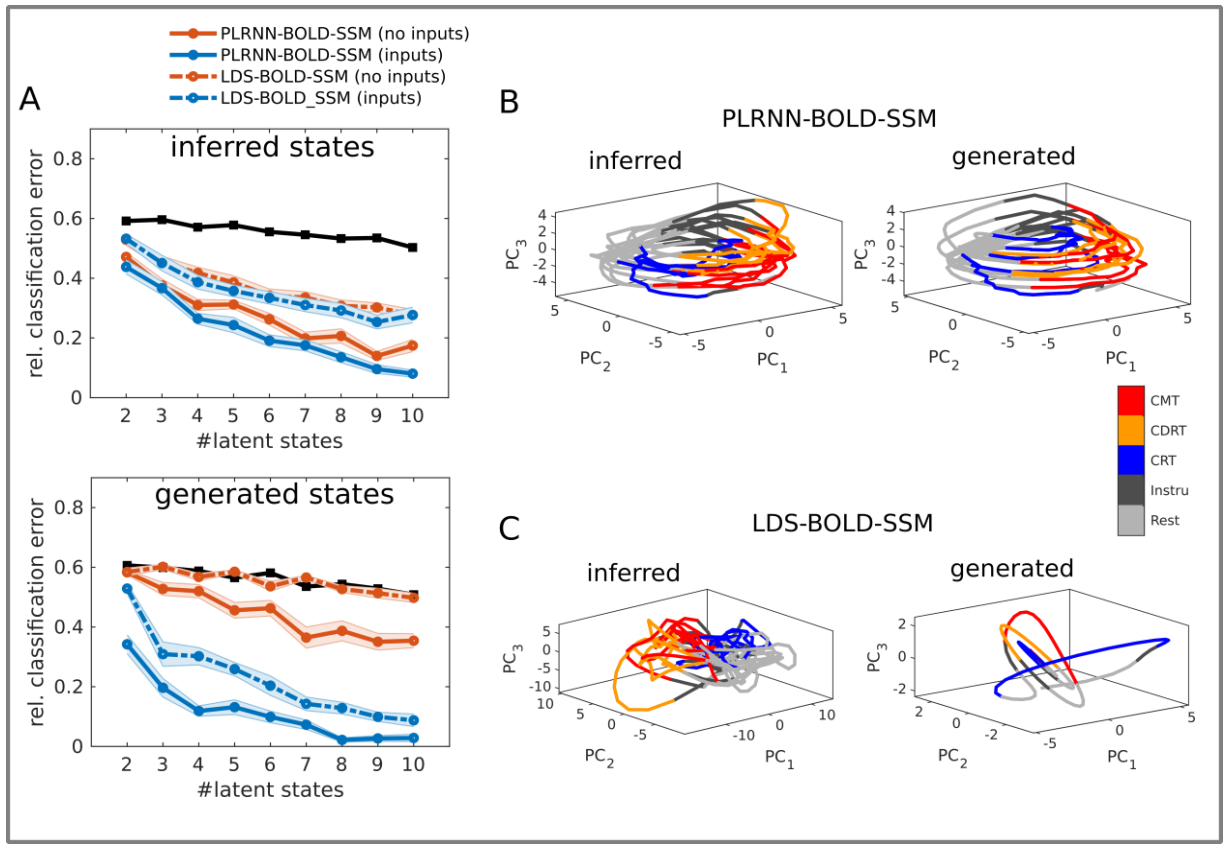

**Fig 7. Decoding task conditions from model trajectories.** A. Relative LDA classification error on different task phases based on the inferred states (top) and freely generated states (bottom) from the PLRNN-BOLD-SSM (solid lines) and LDS-BOLD-SSM (dashed lines), for models including (blue) or excluding (red) stimulus inputs. Black lines indicate classification results for random state permutations. Except for $M=2$, the classification error for the PLRNN-BOLD-SSM based on generated states, drawn from the prior model $p_{gen}(\mathbf{Z})$, is significantly lower than for the permutation bootstraps (all $p<.01$), indicating that the prior dynamics contains task-related information. In contrast, the LDS-BOLD-SSM produced substantially higher discrimination errors for the generated trajectories (which were close to chance level when stimulus information was excluded), and even on the inferred trajectories. Unstable system estimates were removed from analysis. B. Typical example of inferred (left) and generated (right) state space trajectories from a PLRNN-BOLD-SSM, projected down to the first 3 principle components for visualization purposes, color-coded according to task phases (see legend). C. Same as in B for example from trained LDS-BOLD-SSM. The simulated (generated) states usually converged to a fixed point in this case.

To ensure that the retrieved dynamics did not simply capture data variation related to background fluctuations in blood flow (or other systematic effects of no interest), we examined whether the generated trajectories carried task-specific information. For this purpose, we assessed how well we could classify the three experimental tasks (which demanded distinct cognitive processes) via linear discriminant analysis (LDA) based on the *generated* (through the prior model) latent state trajectories. (We exclusively focused on classifying task phases, as these were pseudo-randomized across subjects, while 'resting' and 'instruction' phases occurred at fixed times, and we wanted to prevent significant classification differences which may occur either due to a fixed temporal order, or due to differences in presentation of experimental inputs during resting/instruction vs. proper task phases.) Fig 7A shows the relative classification error obtained when classifying the three tasks by the generated trajectories (bottom) as compared to that from the directly inferred trajectories (top), and to bootstrap permutations of these trajectories (black solid lines).

Overall, for $M>2$ latent states, generated trajectories significantly reduced the relative classification error, even in the absence of any external stimulus information, suggesting that distinct cognitive processes were associated with distinct regions in the latent space, and that this cognitive aspect was captured by the PLRNN-BOLD-SSM prior model (see also Fig 7B for an example of a generated state space for a sample subject, and Fig 8). As observed for the ahead-prediction error above, performance improved with increasing latent state dimensionality. While adding dimensions will boost LDA classifications in general, as it becomes easier to find well separating linear discriminant surfaces in higher dimensions, we did not observe as strong a reduction in classification error for the permutation bootstraps, suggesting that at least part of the observed improvement was related to better reconstruction of the underlying dynamics. Of note, models which included external inputs enabled almost perfect classifications with as few as $M=8$ states. These results are not solely attributable to the model receiving external inputs, as these did not differentiate between cognitive tasks (i.e., number and type of inputs were the same for all tasks, see Methods sect. 'Experimental paradigm').

This is further supported by the observation that the LDS-BOLD-SSM produced much higher classification errors than the PLRNN-BOLD-SSM when either external inputs were present or absent (Fig 7A, dashed lines). Hence, not only does the LDS fail to capture the underlying dynamics and fares worse in ahead predictions (cf. Fig 6D,E), but it also seems to contain less information about the actual task structure, even in the inferred trajectories. This was particularly evident in the situation where trajectories were simulated (generated) and information about external stimuli was not provided to the models, where LDS-BOLD-SSM-based classification performance was close to chance level across all latent state dimensionalities (Fig 7A bottom, red dashed line), consistent with the fact that simulated LDS quickly converged to fixed points (cf. Fig 7C).

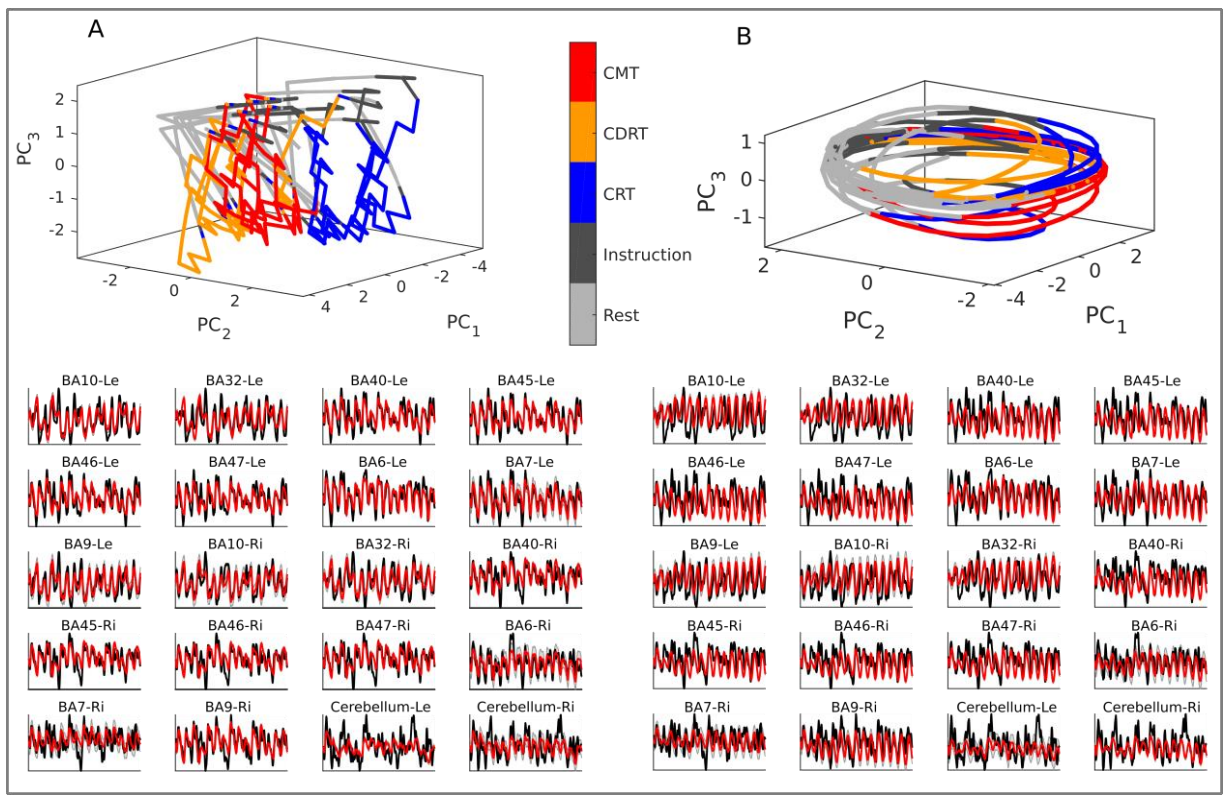

**Fig 8. Exemplary DS reconstruction in a sample subject.** A. Top: Latent trajectories generated by the prior model projected down to the first 3 principle components for visualization purposes in a model including external inputs and *M=6* latent states. Task separation is clearly visible in the generated state space (color-coded as in the legend), i.e. different cognitive demands are associated with different regions of state space (hard step-like changes in state are caused by the external inputs). Bottom: Observed time series (black) and their predictions based on the *generated* trajectories (red, with 90% CI in grey) for the same subject. B. Same as A for the same subject in a PLRNN without external inputs. *BA= Brodmann area, Le/Re=left/right, CRT= choice reaction task, CDRT=continuous delayed response task, CMT=continuous matching task.

Lastly, we observed that trained PLRNN-BOLD-SSMs in many cases produced interesting nonlinear dynamics, including stable limit cycles, chaotic attractors, and multi-stability between various attractor objects (Fig 9). This indicates that the fMRI data may indeed harbor interesting dynamical structure that one would not have been able to reveal with linear state space models like classical DCMs, at least not within the retrieved system of equations (as argued above, the inferred posterior $p(\mathbf{Z}|\mathbf{X})$ may still reflect this structure, but the model itself would not reproduce it).

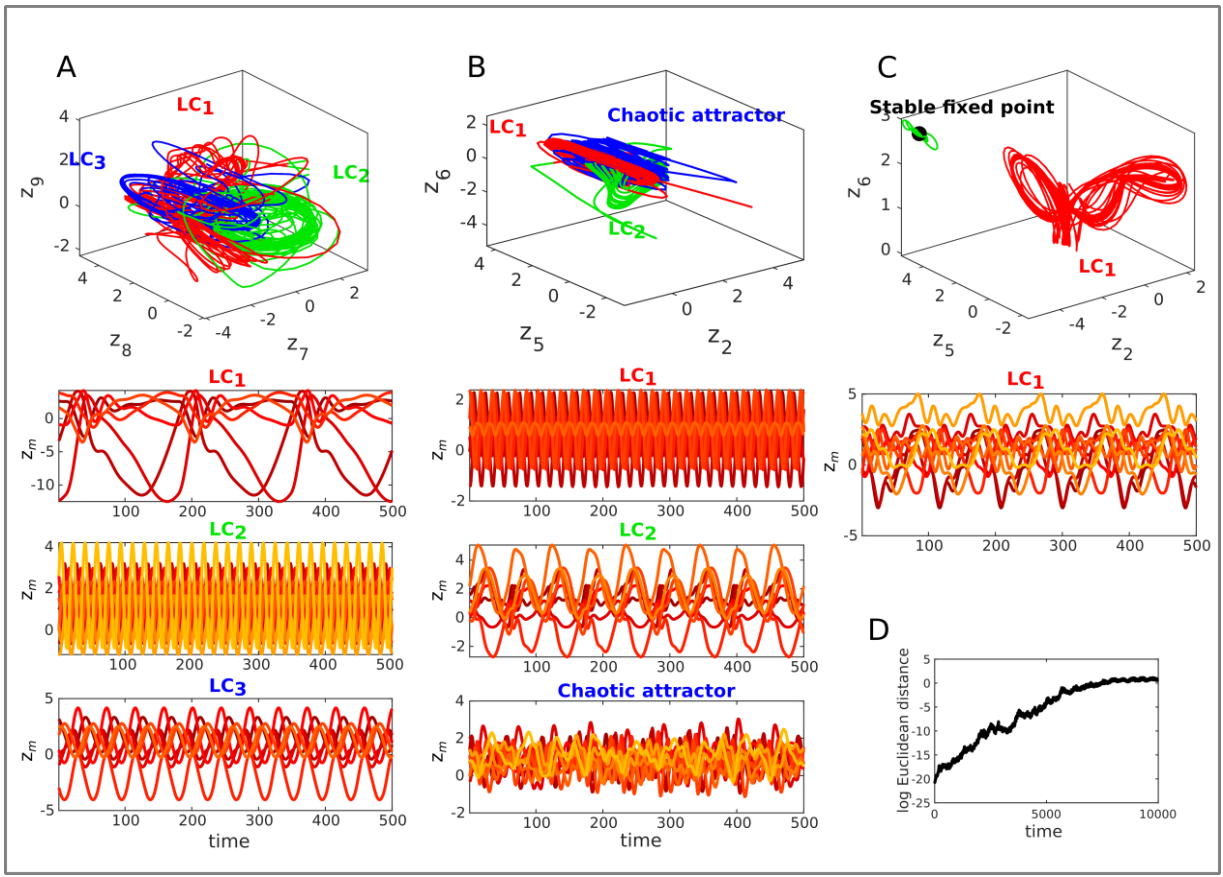

**Fig 9. Examples of highly nonlinear phenomena extracted from fMRI data (in systems with *M=10* states, no external inputs).** A. PLRNN-BOLD-SSM with 3 stable limit cycles (LC) estimated from one subject (top: subspace of state space for 3 selected states; bottom: time graphs). B. PLRNN with 2 stable limit cycles and one chaotic attractor, estimated from another subject. C. PLRNN with one stable limit cycle and one stable fixed point. D. Increase in average (log Euclidean) distance between initially infinitesimally close trajectories with time for chaotic attractor in B. (In A and B states diverging towards $-\infty$ were removed, as by virtue of the ReLU transformation they would not affect the other states and hence overall dynamics).

Furthermore, some of this structure clearly appeared to be linked to task properties: A power spectral analysis of time series generated by the trained PLRNNs revealed that the oscillations exhibited by these models had dominant periods in the same range as the durations of different task phases, as well as periods on the order of the duration of all three different tasks which were delivered in a repetitive manner (Fig 10A). Hence the PLRNN-BOLD-SSM has captured the periodic nature of the experimental design and associated cognitive demands within its limit cycle behavior, even when it was provided with no other source of information than the recorded BOLD activity itself (Fig 10A, left). Moreover, it appeared that the total number of stable objects and unstable fixed points in state space was related to task performance, with better performance (in terms of % correct choices) associated with a lower number of stable but higher number of unstable objects in the CMT (Fig 10B; significant 2-way interaction 'performance-level [low, high] x object-stability [stable, unstable]', $F(1,24)=5.277$, $p=.031$, with performance level based on median split according to % correct choices). This result makes sense from a dynamical systems perspective [e.g. 8, 9, 56], as a lower number of stable objects but large number of unstable fixed points tends to imply a richer and more complex system dynamics which may be associated with better and more flexible cognitive performance. Although such potential links will certainly need to be worked out in much more detail in future studies with potentially more purpose-tailored task designs, these observations illustrate the new possibilities for analyzing links between system dynamics and computational properties provided by our approach, and the new types of questions about neural systems one may be able to ask.

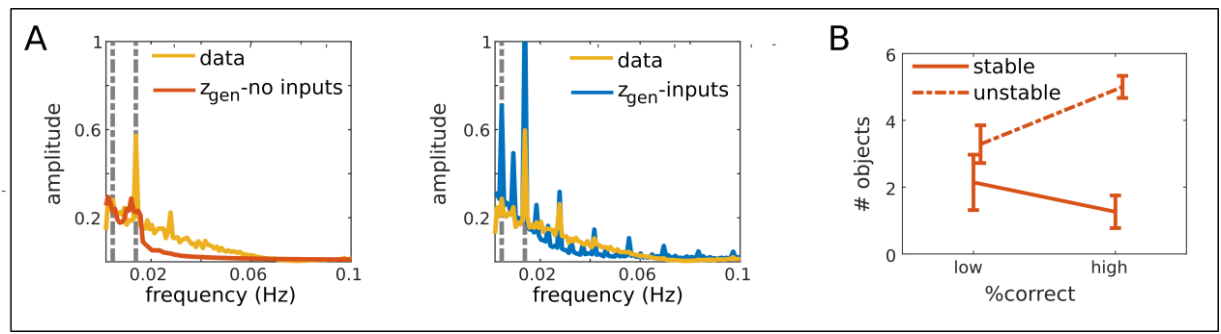

**Fig 10. Links between properties of system dynamics captured by the PLRNN-BOLD-SSM and behavioral task performance.** A. Average power spectra for PLRNN-generated time series when external inputs were excluded (left) and included (right), and for the original BOLD traces (yellow). $M=9$ latent states were used in this analysis, as at this $M$ the number of stable and unstable objects appeared to roughly plateau. The left grey line marks the frequency of one entire task sequence cycle (3·72s=216s=.0046Hz) and the right grey line the frequency of one task and resting block (36s+36s=72s=.0139 Hz). The peaks in the power spectra of the model-generated time series at these points indicate that the PLRNN has captured the periodic reoccurrence of single task blocks as well as that of the whole task block sequence in its limit cycle activity. B. Relation of the number of stable and unstable dynamical objects (see Methods) to behavioral performance for models without external inputs ($M=9$). Low and high performance groups were formed according to median splits over correct responses during the CMT. A repeated measures ANOVA with between-subject factor 'performance' ('low' vs. 'high' percentage of correct responses) and within-subject factor 'stability' ('stable' vs. 'unstable' objects) revealed a significant 'performance x stability' interaction ($F(1,24)=5.28$, $p=.031$). We focused on the CMT for this analysis since for the other two tasks performance was close to a ceiling effect (although results still hold when averaging across tasks, $p=.012$).

**Discussion**

Theories about neural computation and information processing are often formulated in terms of nonlinear DS models, i.e. in terms of attractor states, transitions among these, or transient dynamics still under the influence of attractors or other salient geometrical properties of the state space [4, 9,

57]. Given the success of DS theory in neuroscience, and the recent surge in interest in reconstructing trajectory flows and state spaces from experimental recordings [23, 58-61], methodological tools which would return not only state space representations, but actually a model of the governing equations, would be of great benefit. Here we suggested a novel algorithm within an SSM framework that specifically forces the latent model, represented by a PLRNN, to capture the underlying dynamics in its *intrinsic* behavior, such that it can produce on its own time series of 'fake observations' that closely match the real ones. We also evaluated a measure, the KL divergence defined across state space (not time) between the inferred (posterior) and intrinsically generated (prior) distribution of latent states, which would give us a quantitative sense of how well the underlying DS has been captured in empirical situations where no ground truth is available. Finally, given that fMRI is the most common non-invasive technique to study human cognition in health and psychiatric illness, we derived a new observation model specifically for fMRI data that takes the HRF into account. Using this, we demonstrated that our approach could recover nonlinear dynamics and trajectory flows from human fMRI recordings that were related to task structure and behavioral performance in a working memory paradigm. This, to our knowledge, has not been shown before.

**Choice of model formalism and nonlinearity**

Our major goal here was to establish an efficient methodological approach for recovering dynamical systems from empirical data in a truly generative sense, i.e. such that the trained models exhibit an intrinsic, standalone dynamics that mimics the underlying dynamics of the unknown real system, and to provide a specific measure based on attractor geometries for how well this aim has been achieved. We chose RNNs for the latent model because they are universal approximators of dynamical systems [26-28] and can emulate any Turing machine [62]. Just like the computations performed by a Turing machine can be implemented in many different substrates and algorithmic environments [see, e.g., discussion in 63], the same nonlinear dynamical system and behavior can be implemented in numerous different ways [e.g. 62]. Note, for instance, that the PLRNN can reproduce the chaotic Lorenz attractor although its set of equations is quite different from the original Lorenz equations. Hence, from a pure dynamical systems perspective, the functional form of the nonlinear model, and how close it is to biology, may be largely irrelevant as long as it is powerful enough to approximate any kind of dynamics sufficiently well, i.e. has the required representational expressiveness.

Nevertheless, we would like to repeat that our PLRNN does in fact have the mathematical form of a typical neural rate model as indicated in the first Results section [e.g. 37, 38], and that its ReLU nonlinearity compares quite well to I/O functions of cortical pyramidal cells within the physiologically relevant regime [39, 64, 65], making the model neuronally directly interpretable in principle.

The major reason for settling on a ReLU nonlinearity was, however, that it allows for highly efficient optimization approaches, which also made ReLUs the de-facto standard in modern deep learning applications [44]. In our case, the ReLUs are centerpiece to an efficient fixed-point-iteration-type algorithm for the E-step and enable to compute most expectations analytically and fast (see Methods 'State Estimation'). We believe that this efficiency of optimization, assuring that, in probability, we achieve better approximations to the underlying (biological or physical) system, is more important for capturing biology than the precise functional form of the latent model.

Although this was not a goal here, we further would like to point out that of course also task-specific coupling matrices **W** could be estimated, with subsets of latent states strictly assigned to only certain brain regions (via restrictions on **B**, eqns. 2-3). From a DS perspective, however, one might rather want to think about the same DS (with same parameters) producing different types of tasks (e.g. Yang et al., 2019), where the different tasks are more reflected by different local dynamics in possibly different regions of state space (cf. Fig. 7B) rather than by differences in coupling parameters. Finally, we remark that while one may hope that reconstructing the underlying dynamical system involves a dimensionality reduction ($M < N$), i.e. that the effective dynamics lives in a (much) lower-dimensional space than occupied by the observed measurements, this may not necessarily be the case and there may be situations where the latent space rather has to be expanded ($M > N$) if only relatively few measurement channels are available.

**Comparison to other approaches for identifying dynamical systems**

The 'classical' technique for reconstructing attractor dynamics from experimental time series is delay embedding, based on the delay embedding theorems by Takens (49) and Sauer, Yorke (50). It has been used to disentangle task-related trajectory flows and attractor-like properties in experimentally assessed neuronal time series [22, 23]. However, as a completely non-parametric technique, delay embedding will not give a complete picture of the system's flow field, nor access to the governing equations. Linear dynamical systems, coupled to Gaussian or Poisson observation equations [16, 18, 19], and related approaches like GPFA [20], are quite popular in neurophysiology for obtaining smoothed trajectories and state spaces, but – due to their linear latent dynamics – are inherently unsuitable for reconstructing the underlying DS itself in most cases (as explained above, they may still yield a good approximation to the posterior $p(\mathbf{Z}|\mathbf{X})$, thus still useful, but they would fail to capture the generative dynamics itself as explicitly shown in Fig 7). In consequence, unlike the PLRNN-based models, LDS models were not able to pick up the nonlinear structure present in the BOLD signals in their generative dynamics (but mostly converged to simple fixed points), and probably as a result thereof produced worse forward predictions and contained less information about the cognitive tasks than the PLRNN.

To our knowledge, Roweis and Ghahramani (30), and somewhat later Yu, Afshar (29), were among the first to suggest an RNN for the latent model in order to reconstruct dynamics. These earlier contributions still focused more on in the inferred space $p(\mathbf{Z}|\mathbf{X})$, rather than on the fully generative capabilities of their models (at least were these not systematically analyzed), perhaps partly due to the fact that numerically less stable and efficient inference methods like the extended Kalman filter were employed at the time. Very recent work by Zhao and Park (35) built on the radial basis function networks suggested by Roweis and Ghahramani (30) for the latent model, and combined it with variational inference. They showed ahead predictions of their model for up to 1000 time steps. Similarly, Pandarinath, O'Shea (36) recently proposed a sequential variational auto-encoder framework for inferring dynamics from neural recordings (although here as well the focus was more on the posterior encoding in the latent states, and on inference of initial conditions and perturbations).

Both these models, however, are fairly complex and not directly interpretable in neural terms, and, moreover, hard to analyze with respect to their intrinsic dynamics.

The PLRNN framework offers several distinct advantages compared to other approaches: The equations have a fairly direct neural interpretation [31], in fact have the general form of neural rate equations that have been used to model various neural and cognitive phenomena [37, 38], and – due to their piecewise-linear structure – can also be easily translated into an equivalent *continuous-time* neural rate model [see 66]. Dynamical phenomena can be analyzed more easily in PLRNNs than in other frameworks, e.g. fixed points and their stability can be determined analytically [31]. Furthermore, ReLU-type activation functions appear to be a quite good approximation to the I/O-functions of many neocortical cell types [39, 64], and, besides, are almost the default now in deep networks due to their favorable properties in optimization [44], a feature our iterative state inference algorithm exploits as well. Finally, in contrast to most previous approaches, here we demonstrated that the prior PLRNN model on its own, after training, can produce the same attractor dynamics in state space as the true DS.

In the physics literature, several other methods based on reservoir computing [67], RNNs formed from feedforward networks trained directly on the flow field [see also 26, 28], or LASSO regression combined with polynomial basis expansions [68], have recently been discussed for identifying DS. Process noise is usually not included in these models, i.e. the latent dynamics is deterministic, which entails the risk that noise in the process is wrongly attributed to deterministic aspects of the dynamics. While some of these methods required hundreds of hidden states and millions of samples to reconstruct the van der Pol or Lorenz attractors [28], we found that as few as just eight latent states and a single time series of length 1000, within the range of typical fMRI data, can be sufficient for the PLRNN-SSM to rebuild the chaotic Lorenz attractor, another tremendous advantage in empirical settings.

**Applications in fMRI research and beyond**

In this contribution, we have derived a new observation model for fMRI that accounts for the HRF filtering of the BOLD signal. The HRF implies that current observations do not depend only on the system's current state (the common assumption in SSMs), but on a sequence of previous states, a situation handled relatively seamlessly by our PLRNN-SSM inference algorithm. fMRI is still the most common recording technique for monitoring brain function during cognitive and emotional processing in healthy and psychiatric subjects. Huge data bases have been compiled in large cohort studies over the past decade or so (e.g., the German National Cohort Study initiated by the Helmholtz association: https://www.helmholtz.de/en/research_infrastructures/national_cohort_study/; see also Collins and Varmus (69)) as a reference for monitoring and assessing neurological and psychiatric dysfunction. Although other noninvasive recording techniques with finer temporal resolution, like MEG/ EEG, may be more suitable for addressing questions about the DS basis of cognition, clinical research cannot afford to discard this large body of medically relevant data.

On the other hand, important hypotheses about the neural underpinnings of psychiatric conditions like schizophrenia, attention deficit hyperactivity disorder, or depression, have been formulated in terms of altered system dynamics [see 70 for a recent review]. For instance, based on physiological single unit and synapse data combined with biophysical network models on dopamine modulation in prefrontal cortex, it has been suggested that a dysregulated dopamine system by overly 'deepening' cortical attractor landscapes may inhibit transitions among states, and thereby cause some of the (cognitive) symptoms in schizophrenia [71]. This proposal has been supported by a number of neurophysiological and neuropsychological observations [e.g. 23, 72], but a direct experimental evaluation of the specific changes in attractor basins in schizophrenia is still lacking. Tools like the one proposed here could be applied to directly test these types of hypotheses in human subjects recorded with fMRI. More generally, however, an extensive literature suggests that dynamical properties assessed from fMRI predict psychopathological conditions [e.g. 73, 74, 75], where the methodological framework proposed here could help to better understand the underlying dynamics and define targets for intervention (e.g. in the context of neurofeedback).

Beyond fMRI, most neuroimaging techniques, including, e.g., calcium imaging [76] or imaging by voltage-sensitive dyes [77] in neural tissue, involve some form of filtering that has to be taken into account when the goal is to capture underlying dynamical processes (like neural interactions) that evolve at a faster time scale. Through introduction of a filtering observation model (eq. 3), the present paper establishes a framework for inferring nonlinear dynamics in such situations where the measurement technique involves low- or band-pass-filtering of the process of interest. More generally, while we chose fMRI data here as our applicational example, we emphasize that our methodological framework is generic and could ultimately be applied to any other recording modality, like EEG, MEG, multiple single-unit data, or time series from mobile sensors, ecological momentary assessments [78], or electronic health records, for instance, by simply swapping the observation model eq. 2/3.

**Open issues and outlook**

There is room for improvement in both our training algorithm and the measures used to evaluate its success in empirical situations. Our stepwise training algorithm was devised based on an intuitive heuristic, namely that by shifting the workload for fitting the observations onto the latent model and gradually increasing the requirements for its temporal consistency, a better representation of the unobserved system dynamics could be achieved. We could show that this was indeed the case when compared to a bootstrap (random) sample of models trained in the 'standard' way, and that our procedure seemed to work in general, but a more systematic theoretical derivation and testing of alternative schemes and explicitly designed optimization criteria (directly utilizing eq. 10) would certainly be desirable in future work.

We also find it important that in testing the performance of different reconstruction algorithms not only 'good examples' that prove the basic concept ('my algorithm works') are shown, but a more thorough quantitative statistical evaluation of precisely how well it performed in what percentage of cases is provided, like the one attempted here (Fig 4). For applications to empirical data, for which we do not know the ground truth, an open issue is how we could best quantify how much confidence we could have in the reconstructed stochastic equations of motion. Cross-validation and out-of-sample prediction errors provide a guidance, but for DS it is less clear in terms of what these should be measured: It is known that for nonlinear systems with complex or chaotic dynamics standard squared-error or likelihood-based measures evaluated along time series are not too useful [e.g. 48], since miniscule differences in initial conditions or noise perturbations may cause quick decorrelation of trajectories even if they come from the very same DS. We therefore decided to compare true and simulated data in terms of probability distributions across state space, arguing that if the observations come from the same attractor or system dynamics they should fill roughly the same volume of state space – this is more along the lines of a DS view which compares dynamical objects in terms of their geometrical or topological equivalence in state space [49-51, 79], rather than the literal overlap among time series. Another corollary of this view is that to establish the equivalence between two DS, it is neither sufficient nor potentially even useful to predict observations just a couple of time steps ahead: In a chaotic noisy system, the prediction horizon is inherently limited to begin with (because of exponential divergence of trajectories), and one has to demonstrate that the 'general type' of long-term behavior in the limit is the same (e.g. a limit cycle of a certain periodicity and order) to claim that two DS are equivalent. We therefore suggested to evaluate performance in terms of completely newly generated ('faked') trajectories that the trained system produces when no longer guided by the actual observations (i.e., the prior $p_{gen}(\mathbf{Z})$ rather than the posterior $p_{inf}(\mathbf{Z}|\mathbf{X})$).

Especially in fMRI, however, the data space is often very high-dimensional (>10³) while at the same time often only a single time series sample of limited length ($T$≤1000) is available, i.e. the **x**-space is very sparse. In these cases we cannot obtain a good approximation of the distribution $p(\mathbf{x})$, as we could for the benchmarks, and hence our original measure is not directly applicable. Hence we reverted to performing the comparison in latent space, between two distributions we do have in principle available, the one constrained by the observations, $p_{inf}(\mathbf{z}|\mathbf{x})$, and the other, $p_{gen}(\mathbf{z})$, obtained from the completely freely running (simulated) system. We argued that if our actual observations **X** reflect the true dynamics well, then states obtained under $p_{inf}(\mathbf{z}|\mathbf{x})$ should be highly likely a priori, i.e. under $p_{gen}(\mathbf{z})$, and hence these distributions should highly overlap. As direct sampling from $p_{inf}(\mathbf{z}|\mathbf{x})$ is difficult and time-consuming, due to degeneracy problems, and the latent space dimensionality may also be prohibitively high, we approximated it by a mixture of Gaussians, which is a reasonable assumption for our ReLU-based RNN model and allows for an efficient analytical approximation to $KL_{\mathbf{z}}$ [80]. More generally, if we are only interested in dynamical equivalence, we may also want to accept translations, rotations, rescaling, and potentially other deformations of the true state space that do not change topological equivalence [49, 50]. Procrustes analysis [81] could be performed to (partly) allow for such transformations (on the other hand, since $p_{gen}(\mathbf{Z})$ and $p_{inf}(\mathbf{Z}|\mathbf{X})$ come from the same underlying model, in our specific case such transformations may neither be necessary, nor necessarily desired).

**Methods**

**Model specification and inference**

The formulation of the state space model for BOLD time series (PLRNN-BOLD-SSM) is given in the Results section. To infer the parameters and latent variables of the model, we used Expectation-Maximization (EM) [41, 82]. The EM algorithm maximizes a lower bound $\mathcal{L}(\boldsymbol{\theta}, q)$ (also called the evidence lower bound, ELBO) of the log-likelihood $\log p(\mathbf{X}|\boldsymbol{\theta})$ given by (see S1 Text sect. 'PLRNN-BOLD-SSM model inference' for full details):

(4) $\log p(\mathbf{X}|\boldsymbol{\theta}) \geq \mathrm{E}_q[\log p(\mathbf{X}, \mathbf{Z}|\boldsymbol{\theta})] + H\big(q(\mathbf{Z}|\mathbf{X})\big) = \log p(\mathbf{X}|\boldsymbol{\theta}) - KL(q(\mathbf{Z}|\mathbf{X}), p_{\boldsymbol{\theta}}(\mathbf{Z}|\mathbf{X})) =: \mathcal{L}(\boldsymbol{\theta}, q)$,

with $q(\mathbf{Z}|\mathbf{X})$ some proposal density over latent states, and $KL(q(\mathbf{Z}|\mathbf{X}), p(\mathbf{Z}|\mathbf{X}))$ the Kullback-Leibler divergence between proposal density $q(\mathbf{Z}|\mathbf{X})$ and true posterior $p(\mathbf{Z}|\mathbf{X})$. This expression can be derived by, e.g., using Jensen's inequality [e.g. 30]. From this we see that the bound becomes exact when proposal density $q(\mathbf{Z}|\mathbf{X})$ exactly matches the true posterior density $p(\mathbf{Z}|\mathbf{X})$ (defined through the latent state model here) which we aim to determine in the E-step (in contrast to variational inference where we assume $q(\mathbf{Z}|\mathbf{X})$ to come from some parameterized family of density functions, in EM we usually try to compute [linear case] or approximate $p(\mathbf{Z}|\mathbf{X})$ directly).

**State estimation (E-Step).** In the E-step we seek $q^* := \arg\max_q \mathcal{L}(\boldsymbol{\theta}^*, q)$ given a current parameter estimate $\boldsymbol{\theta}^*$. Since $\boldsymbol{\theta}^*$ is assumed to be given, this amounts to minimizing the Kullback-Leibler divergence $KL(q(\mathbf{Z}|\mathbf{X}), p(\mathbf{Z}|\mathbf{X}))$. The common procedure for linear-Gaussian models [e.g., Kalman filter-smoother; 83, 84] is equating $q(\mathbf{Z}|\mathbf{X}) = p(\mathbf{Z}|\mathbf{X})$, and then determining the first two moments of the latter for performing the M-step. For the present model $p(\mathbf{Z}|\mathbf{X})$ is a high-dimensional mixture of piecewise Gaussians for which 'explicit' integration (i.e., using tabulated Gaussian integrals) becomes unfeasible for large $T$ and $M$. Typically, however, the piecewise Gaussians will have centers close to the origin [S2 Fig; cf. 31], and hence we resort to solving for the maximum a-posteriori (MAP) estimate of $p(\mathbf{Z}|\mathbf{X})$, expected to be close to $\mathrm{E}[\mathbf{Z}|\mathbf{X}]$ (which is exactly so for a single Gaussian), and instantiate the state covariance matrix with the negative inverse Hessian around this maximizer (e.g. [16]). Essentially, this is a global Gaussian approximation, or a Laplace approximation of the log-likelihood where we approximate $\log p(\mathbf{X}|\boldsymbol{\theta}) \approx \log p_{\boldsymbol{\theta}}(\mathbf{X}|\mathbf{Z}^{\max}) + \log p_{\boldsymbol{\theta}}(\mathbf{Z}^{\max}) - \frac{1}{2}\log|-\mathbf{L}^{\max}| + const.$ using the maximizer $\mathbf{Z}^{\max}$ of $\log p_{\boldsymbol{\theta}}(\mathbf{X}, \mathbf{Z})$ (note that the Hessian $\mathbf{L}^{\max}$ is constant around the maximizer) [17, 85]. Taking this approach, letting $\Omega(t) \subseteq \{1 \ldots M\}$ refer to the set of all indices of units for which $z_{m,t} \leq 0$ and $\mathbf{W}_{\Omega(t)}$ to the matrix $\mathbf{W}$ that has all columns corresponding to indices in $\Omega(t)$ set to 0, the optimization objective in the E-Step may be formulated as:

(5) $\max\{ Q_\Omega^*(\mathbf{Z}) := -\frac{1}{2}(\mathbf{z}_1 - \boldsymbol{\mu}_0 - \mathbf{Cs}_1)^\mathrm{T} \boldsymbol{\Sigma}^{-1}(\mathbf{z}_1 - \boldsymbol{\mu}_0 - \mathbf{Cs}_1)$

$-\frac{1}{2}\sum_{t=2}^{T}(\mathbf{z}_t - (\mathbf{A} + \mathbf{W}_{\Omega(t-1)})\mathbf{z}_{t-1} - \mathbf{h} - \mathbf{Cs}_t)^\mathrm{T} \boldsymbol{\Sigma}^{-1}(\mathbf{z}_t - (\mathbf{A} + \mathbf{W}_{\Omega(t-1)})\mathbf{z}_{t-1} - \mathbf{h} - \mathbf{Cs}_t)$

$-\frac{1}{2}\sum_{t=1}^{T}(\mathbf{x}_t - \mathbf{B}(\mathrm{hrf} * \mathbf{z}_{\tau:t}) - \mathbf{Jr}_t)^\mathrm{T} \boldsymbol{\Gamma}^{-1}(\mathbf{x}_t - \mathbf{B}(\mathrm{hrf} * \mathbf{z}_{\tau:t}) - \mathbf{Jr}_t) + \mathrm{const}\}$

w.r.t. $(\Omega, \mathbf{Z})$ subject to $z_{i,t} \leq 0 \; \forall \; i \in \Omega(t) \land z_{i,t} > 0 \; \forall \; i \notin \Omega(t) \; \forall \; t$.

Let us concatenate all state variables across $m$ and $t$ into one long column vector $\mathbf{z} = (z_{11}, \ldots, z_{M1}, \ldots, z_{1T}, \ldots, z_{MT})^\mathrm{T} \in \mathbb{R}^{MT}$, and likewise arrange all matrices $\mathbf{A}$, $\mathbf{W}_{\Omega(t)}$, and so on, into large $MT \times MT$ block tri-diagonal matrices, and let us further collect all terms quadratic in $\mathbf{z}$, linear in $\mathbf{z}$, or constant (see S1 Text for exact composition of these matrices). Defining $\mathbf{H}$ as the HRF convolution matrix, $\mathbf{d}_\Omega := (\mathrm{I}(z_{11} > 0), \mathrm{I}(z_{21} > 0), \ldots, \mathrm{I}(z_{MT} > 0))^\mathrm{T}$ as an indicator vector with a 1 for all states $z_{m,t} > 0$ and zeros otherwise, and $\mathbf{D}_\Omega := diag(\mathbf{d}_\Omega)$ as the diagonal matrix formed from this vector, one can rewrite the optimization criterion (eq. 5) compactly as

(6) $Q_\Omega^*(\mathbf{Z}) = -\frac{1}{2}[\mathbf{z}^\mathrm{T}(\mathbf{U}_0 + \mathbf{D}_\Omega \mathbf{U}_1 + \mathbf{U}_1^\mathrm{T} \mathbf{D}_\Omega + \mathbf{D}_\Omega \mathbf{U}_2 \mathbf{D}_\Omega + \mathbf{H}^\mathrm{T} \mathbf{U}_3 \mathbf{H})\mathbf{z} - \mathbf{z}^\mathrm{T}(\mathbf{v}_0 + \mathbf{D}_\Omega \mathbf{v}_1 + \mathbf{H}^\mathrm{T} \mathbf{v}_2) -$

$(\mathbf{v}_0 + \mathbf{D}_\Omega \mathbf{v}_1 + \mathbf{H}^\mathrm{T} \mathbf{v}_2)^\mathrm{T} \mathbf{z}] + \mathrm{const}$,

which is a piecewise quadratic function in **z** with solution vectors

$$\mathbf{z}^* = [\mathbf{U}_0 + \mathbf{D}_\Omega \mathbf{U}_1 + \mathbf{U}_1^T \mathbf{D}_\Omega + \mathbf{D}_\Omega \mathbf{U}_2 \mathbf{D}_\Omega + \frac{1}{2}(\mathbf{H}^T \mathbf{U}_3 \mathbf{H} + (\mathbf{H}^T \mathbf{U}_3 \mathbf{H})^T)]^{-1}[\mathbf{v}_0 + \mathbf{D}_\Omega \mathbf{v}_1 + \mathbf{H}^T \mathbf{v}_2],$$

*provided* this solution is consistent with the current set $\Omega$, i.e. is a true solution of eq. 6. For solving this set of piecewise linear equations, we use a simple Newton-type iteration scheme, similar to the one suggested in [86], where we iterate between (1) solving eq. 6 for fixed $\mathbf{d}_\Omega$ and (2) flipping the bits in $\mathbf{d}_\Omega$ inconsistent with the obtained solution to eq. 6, until convergence. Care is taken to avoid getting trapped in cyclic behavior, and a quadratic programming step may be added at the end to obtain the maximum given a fixed index set $\Omega$ [which seemed rarely necessary from our experience; see 31 for details].

Once a solution $\mathbf{z}^*$ with high posterior density has been obtained, the state covariance matrix is approximated locally around this estimate as the inverse negative Hessian

$$\mathbf{V} = [\mathbf{U}_0 + \mathbf{D}_\Omega \mathbf{U}_1 + \mathbf{U}_1^T \mathbf{D}_\Omega + \mathbf{D}_\Omega \mathbf{U}_2 \mathbf{D}_\Omega + \frac{1}{2}(\mathbf{H}^T \mathbf{U}_3 \mathbf{H} + (\mathbf{H}^T \mathbf{U}_3 \mathbf{H})^T)]^{-1}.$$

These state covariance estimates are then used to compute, mostly analytically, the expectations $E[\varphi(\mathbf{z})]$, $E[\mathbf{z}\varphi(\mathbf{z})^T]$, and $E[\varphi(\mathbf{z})\varphi(\mathbf{z})^T]$ required in the M-Step [please see S1 Text and 31 for more details]. This global iterative E-Step scheme is particularly suitable for fMRI applications in which the HRF invokes temporal dependencies between current observations and latent states that reach back in time by several lags (i.e. $\mathbf{x}_t$ does not only depend on $\mathbf{z}_t$, but on a set of previous states $\mathbf{z}_{\tau:t}$). This implies that $p(\mathbf{Z}|\mathbf{X})$ does not factorize as required for the common (unscented or extended) Kalman filter. Although our approach is global, as pointed out by Paninski, Ahmadian (17), efficient schemes for inverting block-tridiagonal matrices still scale linearly in $T$.

**Parameter estimation (M-Step).** In the M-step, parameters are updated by seeking $\boldsymbol{\theta}^* \coloneqq \arg\max_{\boldsymbol{\theta}} \mathcal{L}(\boldsymbol{\theta}, q^*)$ given $q^*$ from the E-step (since $q^*$ is assumed fixed and known in the E-step, note that the entropy over $q$ becomes a constant in eq. 4 and drops out from the maximization). This boils down to a simple linear regression problem given that the ReLU nonlinearities have been resolved within the expectations $E[\varphi(\mathbf{z})]$, $E[\mathbf{z}\varphi(\mathbf{z})^T]$, and $E[\varphi(\mathbf{z})\varphi(\mathbf{z})^T]$, and hence criterion eq. 5 becomes simply quadratic.

We can (analytically) solve for the parameters $\boldsymbol{\theta}_{obs}$ of the observation model and $\boldsymbol{\theta}_{lat}$ of the latent model separately. Because of the off-diagonal structure of **W**, it is most efficient to obtain parameter solutions row-wise for the latent model parameters (i.e., separately for each state $m=1...M$), as spelled out in S1 Text. For the observation model parameters, concatenating matrices **B** and **J** as $\mathbf{Y} = [\mathbf{B}\ \mathbf{J}] \in \mathbb{R}^{N \times (M+P)}$, and concatenating convolved states and nuisance variables in $\mathbf{y}_t \in \mathbb{R}^{M+P}$, one can rewrite the observation equation term in $Q(\boldsymbol{\theta}, \mathbf{Z}) \coloneqq E_q[\log p(\mathbf{X}, \mathbf{Z}|\boldsymbol{\theta})]$ as

(7) $\quad Q_{obs}(\boldsymbol{\theta}_{obs}, \mathbf{Z}) = -\frac{1}{2}\sum_{t=1}^{T} E[(\mathbf{x}_t - \mathbf{Y}\mathbf{y}_t)^T \boldsymbol{\Gamma}^{-1}(\mathbf{x}_t - \mathbf{Y}\mathbf{y}_t)] - \frac{T}{2}\log|\boldsymbol{\Gamma}|$

Differentiating w.r.t. to **Y** and setting to 0 yields

$$\mathbf{Y} = \left(\sum_{t=1}^{T} E[\mathbf{x}_t \mathbf{y}_t^T]\right)\left(\sum_{t=1}^{T} E[\mathbf{y}_t \mathbf{y}_t^T]\right)^{-1}.$$

Defining the sums of cross-products

$$\mathbf{F}_2 := \sum_{t=1}^{T} \mathbf{x}_t \mathbf{x}_t^{\mathrm{T}}, \qquad \mathbf{F}_7 := \sum_{t=1}^{T} \mathbf{x}_t \mathbf{r}_t^{\mathrm{T}}, \qquad \mathbf{F}_8 := \sum_{t=1}^{T} \mathbf{r}_t \mathbf{r}_t^{\mathrm{T}}, \qquad \mathbf{H}_1 := \sum_{t=1}^{T} \mathbf{x}_t \mathrm{E}[(hrf * \mathbf{z}_{\tau:t})^{\mathrm{T}}],$$

$$\mathbf{H}_2 := \sum_{t=1}^{T} \mathbf{r}_t \mathrm{E}[(hrf * \mathbf{z}_{\tau:t})^{\mathrm{T}}], \qquad \mathbf{H}_3 := \sum_{t=1}^{T} \mathrm{E}[(hrf * \mathbf{z}_{\tau:t})(hrf * \mathbf{z}_{\tau:t})^{\mathrm{T}}]$$

we can equivalently express the solution as

$$\mathbf{Y} = [\mathbf{H}_1 \; \mathbf{F}_7] \begin{bmatrix} \mathbf{H}_3 & \mathbf{H}_2^{\mathrm{T}} \\ \mathbf{H}_2 & \mathbf{F}_8 \end{bmatrix}^{-1}, \qquad \mathbf{B} = \mathbf{Y}_{1:N,1:M}, \qquad \mathbf{J} = \mathbf{Y}_{1:N,M+1:M+P}.$$

With these definitions, differentiating eq. 7 w.r.t $\mathbf{\Gamma}$ yields

$$\mathbf{\Gamma} = \frac{1}{T} (\mathbf{F}_2 - \mathbf{H}_1 \mathbf{B}^{\mathrm{T}} - \mathbf{B} \mathbf{H}_1^{\mathrm{T}} + \mathbf{B} \mathbf{H}_3^{\mathrm{T}} \mathbf{B}^{\mathrm{T}} - \mathbf{F}_7 \mathbf{J}^{\mathrm{T}} - \mathbf{J} \mathbf{F}_7^{\mathrm{T}} + \mathbf{B} \mathbf{H}_2^{\mathrm{T}} \mathbf{J}^{\mathrm{T}} + \mathbf{J} \mathbf{H}_2^{\mathrm{T}} \mathbf{B}^{\mathrm{T}} + \mathbf{J} \mathbf{F}_8 \mathbf{J}^{\mathrm{T}}) \circ \mathbf{I}$$

where $\mathbf{I}$ denotes an $N$x$N$ identity matrix. Solutions for the latent state parameters $\mathbf{\theta}_{lat}$ are given in S1 Text. E- and M-steps are then iterated until convergence of the expected joint log-likelihood.

**Stepwise model training procedure**

We introduce here an efficient approach for pushing the latent model to capture the underlying DS that generated the observations. Our approach rests on a step-wise procedure in which we gradually increase the importance of fitting the latent state dynamics as compared to fitting the observations. Since the latent state process and the observation process account for additive terms in the joint log-likelihood (eq. 5), the tradeoff between fitting the dynamics and fitting the observations is regulated by the ratio of the two covariance matrices $\mathbf{\Sigma}$ and $\mathbf{\Gamma}$ (eqns. 1-3,5). Hence, the idea of our training scheme is to begin with fitting the observation model and putting milder constraints on the latent process, using a *linear* latent model for initialization in a first step [or even factor analysis which places no constraints on the temporal relationship among latent states; cf. 30], and then gradually decreasing "$\mathbf{\Sigma}:\mathbf{\Gamma}$" during training to enforce the temporal consistency of the latent model. Furthermore, one may force all burden of fitting the observations completely onto the latent model by fixing $\mathbf{\theta}_{obs}$ from some step onwards. The complete training protocol is outlined in Algorithm-1. For inferring a linear model (LDS-SSM, LDS-BOLD-SSM), the exact same algorithm was used with $\varphi(\mathbf{z}) = \max(\mathbf{z}, 0)$ just replaced by $\varphi(\mathbf{z}) = \mathbf{z}$ in eqns. 1-2.

**Algorithm-1**

> 0) Draw initial parameter estimates $\theta^{(0)} \sim p(\theta)$ from some suitable prior, constraint to $\max |\text{eig}(\mathbf{A} + \mathbf{W})| < 1$ for biasing toward stable models [see also 18].
> 1) Fix $\mathbf{\Sigma} = \mathbf{I}$ and run linear dynamical system (LDS) SSM for initialization → $\theta^{(1)}$
> 2) Fix $\mathbf{\Sigma} = \mathbf{I}$ and run PLRNN-SSM inference → $\theta^{(2)}$
> 3) for $i = 1:3$
>    - Fix $\mathbf{\Sigma} = \text{diag}(10^{-i})$, $\mathbf{B} = \mathbf{B}^{(2)}$; fix $\mathbf{\Gamma} = \mathbf{\Gamma}^{(2)}$ (for fMRI data)
>    - Initialize PLRNN-SSM training with previous estimate $\theta^{(i+1)}$
>    - Run PLRNN-SSM inference → $\theta^{(i+2)}$
> 4) Re-estimate state covariance matrix $\text{Var}(\mathbf{z}_t | \mathbf{x}_{1:T})$ with $\mathbf{\Sigma} = \mathbf{I}$ fixed.

**Reconstruction of benchmark dynamical systems**

We evaluated the performance of our PLRNN-SSM approach (and an LDS-SSM for comparison), on two popular benchmark DS, the Lorenz equations and the van der Pol nonlinear oscillator (vdP). Within some parameter range, the 3-dimensional Lorenz system exhibits a chaotic attractor and the 2-dimensional vdP-system exhibits a limit cycle (see Fig 4 for parameter settings used, system equations, and sample trajectories of the systems). We were interested in solutions where the true system dynamics is not just reflected in the directly inferred posterior distribution $p(\mathbf{Z}|\mathbf{X})$ over the PLRNN states $\{\mathbf{z}_{1:T}\}$ given the actual observations $\{\mathbf{x}_{1:T}\}$, but also in the model's generative or prior distribution $p(\mathbf{Z})$, i.e. whether the once estimated PLRNN when run on its own would produce similar trajectories with the same dynamical properties as the ground truth system.

For evaluation, $n$=100 samples of (standardized) trajectories of length $T$=1000 were drawn from the ground truth systems using Runge-Kutta numerical integration and random initial conditions. PLRNN-SSMs were trained on these sample sets as described above for $M$=5…20 latent states, using eq. 2 for the observations (see also Fig 1). To probe our stepwise training protocol (Algorithm-1), PLRNN-SSM training under this protocol (termed 'PLRNN-SSM-anneal') was compared to simple EM training of the PLRNN-SSM started from random initializations of parameters (termed 'PLRNN-SSM-random'; essentially just step 1 of Algorithm-1 with $\mathbf{\Sigma}$ directly fixed to $10^{-3}$) for $M$={8, 10, 12, 14}.

To quantify how well the true system dynamics was captured by the 'free-running' PLRNN (after training, but unconstrained by the observations), we used the Kullback-Leibler divergence defined across *state space*, i.e. integrating across space, not across time. Similar in spirit to the criteria defined for the classical delay embedding theorems [49-51], our measure therefore assessed the agreement between the original and reconstructed *attractor geometries*. Integrating across time (i.e., computing divergence between time series) is problematic for nonlinear DS, since two time series from the very same chaotic DS usually cannot be expected to overlap very well with even miniscule differences in initial conditions [cf. 48]. For the ground truth benchmark systems, for which we have access to the true distribution $p_{true}(\mathbf{x})$ and the complete state space, this KL divergence can be computed directly in observation space and was defined as

$$(8)\quad KL_{\mathbf{x}}\big(p_{true}(\mathbf{x}), p_{gen}(\mathbf{x}|\mathbf{z})\big) := \int_{\mathbf{x}\in\mathbb{R}^N} p_{true}(\mathbf{x}) \log\frac{p_{true}(\mathbf{x})}{p_{gen}(\mathbf{x}|\mathbf{z})}\, d\mathbf{x},$$

where the integration is performed across **x**-space, and $p_{gen}(\mathbf{x}|\mathbf{z})$ is the distribution across observations generated from PLRNN simulations (i.e., after PLRNN-SSM training, but discarding the original set of time series observations $\mathbf{X}^{obs} = \{\mathbf{x}_{1:T}\}$ used for training). Hence, this measure assesses whether PLRNN-SSM-simulated trajectories in the limit fill the same volume of state space as the true DS trajectories, and in this sense whether the systems' attractor objects are topologically and geometrically 'equivalent'. (As a terminological remark, in the machine learning literature $p_{gen}(\mathbf{x}|\mathbf{z})$ is often called the 'generative' or 'decoding' model, while $p(\mathbf{z}|\mathbf{x})$ or $q(\mathbf{z}|\mathbf{x})$ is sometimes referred to as the 'encoder' or 'recognition' model [e.g. 32, 87]. Here we will, more generally, refer with $p_{gen}(\mathbf{z})$ to the (prior) distribution of latent states generated by the PLRNN *independent of the training observations* $\mathbf{X}^{obs} = \{\mathbf{x}_{1:T}\}$, and with $p_{gen}(\mathbf{x}|\mathbf{z})$ to the distribution of *simulated* observations produced from samples $\mathbf{z}^{gen} \sim p_{gen}(\mathbf{z})$ according to the observation model [eq. 2]).

Practically, we discretized the **x**-space into *K* bins of width Δ**x** and evaluated the probabilities 'empirically' as relative frequencies $\hat{p}^{(k)} = \frac{\#_k}{T}$ by filling the space with trajectories ($T = 100{,}000$) sampled from the true DS and trained PLRNNs (here we used Δ**x** = 1 across a range $x_n \in [-4\ 4]$ for standardized variables, but smaller bin sizes yielded qualitatively similar results, see S3 Fig). To avoid $\hat{p}_k(\mathbf{x}|\mathbf{z}) = 0$ for the generative model, where the KL divergence is not defined, we further adjusted this relative frequency to $\hat{p}^{(k)} = \frac{n(k)+\alpha}{T+\alpha K}$, with $\alpha = 10^{-6}$, also known as Laplace or additive smoothing [88] such that eq. 8 becomes

$$(9)\quad KL_{\mathbf{x}}\big(p_{true}(\mathbf{x}), p_{gen}(\mathbf{x}|\mathbf{z})\big) \approx \sum_{k=1}^{K} \hat{p}^{(k)}_{true}(\mathbf{x}) \log\left(\frac{\hat{p}^{(k)}_{true}(\mathbf{x})}{\hat{p}^{(k)}_{gen}(\mathbf{x}|\mathbf{z})}\right).$$

Lastly, to obtain an interpretable measure between 0 and 1, we normalized the KL divergence (termed $\widetilde{KL}_{\mathbf{x}}$) by dividing it by the expected maximum deviation. $\widetilde{KL}_{\mathbf{x}}$ and the expected joint log-likelihood were compared between PLRNN-SSM-anneal and PLRNN-SSM-random via independent *t*-tests. For these analyses, all unstable system estimates were removed (≈14%). Furthermore, strong outliers with joint log-likelihood values < -1000 (which occurred only for PLRNN-SSM-random in ≈3.8% of cases) were removed.

A standard measure of chaoticity in nonlinear DS is the maximal Lyapunov exponent [24]. We thus also assessed how well our KL measure correlated with the deviation in Lyapunov exponents between true and estimated systems. The Lyapunov exponent was assessed numerically by a linear regression fit to the initial slope of the log-Euclidean distance $\log d_{\Delta t}(\mathbf{X}^{(1)}, \mathbf{X}^{(2)})$ between initially close ($d_0 < 10^{-10}$) trajectories $\mathbf{X}^{(1)}$ and $\mathbf{X}^{(2)}$ as a function of time lag $\Delta t$, up to the point in the curve where a plateau indicating the full extent of the attractor object has been reached. For the van der Pol nonlinear (non-chaotic) oscillator, the agreement in the power spectra between the true and generated systems is more informative as a measure of how well the system dynamics has been captured (the maximum Lyapunov exponent for a stable limit cycle is 0), which was simply assessed by the average Pearson correlation.

**Reconstruction of dynamical systems from experimental data**

**Ethics statement.** The human data analyzed here has been collected within a study approved by the local ethics committee of the University of Giessen, School of Medicine, and written informed consent was obtained from each participant prior to enrollment (AZ 63/08).

**Experimental paradigm.** The experimental paradigm assessed three cognitive tasks, two working memory (WM) n-back tasks - the continuous delayed response task (CDRT), and the continuous matching task (CMT) - and a choice reaction task (CRT), which served as 0-back control task. In all tasks, subjects were presented with a sequence of stimuli, and they had to respond to each stimulus (a triangle or a square) according to the task instruction. While in the CDRT participants were asked to indicate which stimulus was presented last, the CMT required participants to compare the current to the last stimulus and indicate whether they were the same or different [89]. In the CRT, participants had to simply indicate the current stimulus, and WM was not required. The paradigm is known to robustly activate the WM network. Each task was preceded by a resting period and an instruction phase. Tasks only differed w.r.t. the instruction phase, otherwise participants were faced with the same stimulus sequence, presented on a central screen at variable inter-stimulus intervals.

**Data acquisition and preprocessing.** Exact details on fMRI data acquisition and preprocessing, as well as information on the sample and consent of study participation can be found in [54]. In brief, 26 healthy subjects participated in the study, undergoing the experimental paradigm in a 1.5 GE Scanner. From these data, we chose to preselect voxel time series known to be relevant to the n-back task, as identified by a previous meta-analysis [55]. This included the following Brodmann areas (BA): BA6 (supplementary motor), BA32 (anterior cingulate), BA46, BA9 (dorsolateral prefrontal cortices), BA45, BA47 (ventromedial prefrontal cortices), BA10 (orbitofrontal cortex), BA7, BA40 (parietal cortices), as well as the medial cerebellum. From each of these areas we extracted the first principle component. Given 10 bilateral regions, this amounted to extracting 20 voxel time series from each participant. Time series were mean centered, and mildly temporally smoothed by convolution with a Gaussian filter ($\sigma^2=1$).

For each individual, the 20 extracted time series were entered as experimental observations **X** along with 6 nuisance predictors **R** (related to movement vectors obtained from the SPM realignment preprocessing procedure) [54] to the PLRNN-BOLD-SSM inference procedure. The LDS-BOLD-SSM was set up the same way (see above), while for the PLRNN fit directly on the observations we set $M=N$ and restricted **B** (eq. 3) to be a diagonal matrix, thus creating a strict 1:1 mapping between 'latent states' and observations. This essentially converts the model into a nonlinear auto-regressive-type model formulated directly on the observations and eliminates the degrees of freedom associated with true latent states.

All models were estimated both including and excluding experimental inputs. For the inclusion condition, experimental inputs **S** were defined as binary 'design' vectors of length $K=5$. The first two entries contained 1's for the presentation of the two stimulus types ('triangle' or 'square'), and the last 3 entries indicated by 1's the instruction phases of the three tasks; all other entries were set to 0. Note that during the actual task phases (following the instruction phases) the inference algorithm therefore

(like the real subjects) received only information about the presented stimuli but not about the task phase itself. Models were estimated with $L_2$ regularization and regularization factor $\lambda=50$.

**Assessment of dynamical objects**. For the PLRNN as formulated in eq. 1, fixed points $\mathbf{z}^*$ can be determined analytically by assessing the solutions $\mathbf{z}^* = (\mathbf{I}_M - \mathbf{A} - \mathbf{W}\mathbf{D}_\Omega)^{-1}\mathbf{h}$ for all $2^M$ configurations of the matrix $\mathbf{D}_\Omega$ as defined above. A fixed point $\mathbf{z}_\Omega^*$ for which the maximum absolute eigenvalue of the corresponding matrix $\mathbf{A} + \mathbf{W}\mathbf{D}_\Omega$ is larger than 1 is unstable, and (neutrally) stable otherwise. Limit cycles and chaotic attractors were assessed by running each system from 100 random initial conditions for *T=5000* time steps. If the system converged to a stable pattern in this limit, it was considered a chaotic attractor if the log-Euclidean distance between two trajectories started from infinitesimally close initial conditions was growing over time (i.e. had a positive slope, see last section on Lyapunov exponents), and a stable limit cycle otherwise (although for the results presented here this distinction does not play a role). The number of stable objects was then determined as the total number of stable fixed points, limit cycles, and chaotic attractors counted this way.

**Reconstruction measures.** In the case of experimental data, in which the ground truth DS is not known, we do not have access to the data generating distribution $p_{true}(\mathbf{X})$, nor to the complete state space in general. We therefore used as a proxy for eq. 9 the Kullback-Leibler divergence between the distribution over latent states obtained by sampling from the data-*unconstrained* prior $p_{gen}(\mathbf{z})$ and the data-*constrained* (i.e., inferred) posterior distribution $p_{inf}(\mathbf{z}|\mathbf{x})$, arguing that the former should match closely with the latter if the actually observed $\mathbf{x}$ represent the underlying DS well (see Results section; also note that the $\mathbf{z}$-space is always complete by model definition, at least in the autonomous case). We again take the KL divergence across the system's state space (not time):

$$(10)\quad KL_\mathbf{z}\big(p_{inf}(\mathbf{z}|\mathbf{x}), p_{gen}(\mathbf{z})\big) = \int_{\mathbf{z}\in\mathbb{R}^M} p_{inf}(\mathbf{z}|\mathbf{x}) \log \frac{p_{inf}(\mathbf{z}|\mathbf{x})}{p_{gen}(\mathbf{z})} d\mathbf{z}\,.$$

To evaluate this integral, sampling from $p_{inf}(\mathbf{z}|\mathbf{x})$, however, is difficult because of the known degeneracy problems with particle filters or other numerical samplers in high dimensions [90, 91]. We therefore approximated both $p_{inf}(\mathbf{z}|\mathbf{x})$ and $p_{gen}(\mathbf{z})$ as Gaussian mixtures across trajectory times, i.e. with $p_{inf}(\mathbf{z}|\mathbf{x}) \approx \frac{1}{T}\sum_{t=1}^T p(\mathbf{z}_t|\mathbf{x}_{1:T})$ and $p_{gen}(\mathbf{z}) \approx \frac{1}{T}\sum_{t=1}^T p(\mathbf{z}_t|\mathbf{z}_{t-1})$, which is reasonable given that the PLRNN distribution is a mixture of piecewise Gaussians (see above). Just as in eqns. 8-9 above, probabilities are therefore evaluated *in space across all time points*. The mean and covariance of $p(\mathbf{z}_t|\mathbf{x}_{1:T})$ and $p(\mathbf{z}_t|\mathbf{z}_{t-1})$ were obtained by marginalizing over the multivariate distributions $p(\mathbf{Z}|\mathbf{X})$ and $p_{gen}(\mathbf{Z})$, respectively, yielding $\mathrm{E}[\mathbf{z}_t|\mathbf{x}_{1:T}], \mathrm{E}[\mathbf{z}_t|\mathbf{z}_{t-1}]$, and covariance matrices $\mathrm{Var}(\mathbf{z}_t|\mathbf{x}_{1:T})$ and $\mathrm{Var}(\mathbf{z}_t|\mathbf{z}_{t-1})$. Note that the covariance matrix of $p(\mathbf{Z}|\mathbf{X})$ was re-estimated at the end of the full training procedure with the process noise matrix $\mathbf{\Sigma}$ set to the identity (i.e., to the last value it had before $\mathbf{\Gamma}$ was fixed qua Algorithm-1). Diagonal elements of the covariance matrix of $p(\mathbf{Z}|\mathbf{X})$ were further restricted to a minimum value of 1 (some lower bound on the variance turned out to be necessary to make $KL_\mathbf{z}$ well defined almost everywhere).

Finally, the integral in eq. 10 was numerically approximated through Monte Carlo (MC) sampling [80] using *n*=500,000 samples:

$$(11) \quad KL_{\mathbf{z}}\big(p_{inf}(\mathbf{z}|\mathbf{x}), p_{gen}(\mathbf{z})\big) \approx \frac{1}{n}\sum_{i=1}^{n} \log \frac{\sum_{t=1}^{T} p(\mathbf{z}^{(i)}|\mathbf{x}_{1:T})/T}{\sum_{l=1}^{L} p(\mathbf{z}^{(i)}|\mathbf{z}_{l-1})/L}.$$

For high-dimensional latent spaces, (asymptotically unbiased) approximation through MC sampling becomes computationally inefficient or unfeasible. For these cases, Hershey and Olson (2007) suggest a variational approximation to the integral in eq. 10 which we found to be in almost exact agreement with the results obtained through MC sampling:

$$(12) \quad KL_{\mathbf{z}}^{(variational)}\big(p_{inf}(\mathbf{z}|\mathbf{x}), p_{gen}(\mathbf{z})\big) \approx \frac{1}{T}\sum_{t=1}^{T} \log \frac{\sum_{j=1}^{T} e^{-KL(p_{inf}(\mathbf{Z}_t|\mathbf{X}_{1:T}), p_{inf}(\mathbf{Z}_j|\mathbf{X}_{1:T}))}}{\sum_{k=1}^{T} e^{-KL(p_{inf}(\mathbf{Z}_t|\mathbf{X}_{1:T}), p_{gen}(\mathbf{Z}_k|\mathbf{Z}_{k-1}))}},$$

where the terms in the exponentials refer to KL divergences between pairs of Gaussians, for which an analytical expression exists.

We normalized this measure by dividing by the KL divergence between $p_{inf}(\mathbf{z}|\mathbf{x})$ and a reference distribution $p_{ref}(\mathbf{z})$ which was simply given by the temporal average across state expectations and variances along trajectories of the prior $p_{gen}(\mathbf{Z})$ (i.e., by one big Gaussian in an, on average, similar location as the Gaussian mixture components, but eliminating information about spatial trajectory flows). (Note that we may rewrite the evidence lower bound as $\mathcal{L}(\mathbf{\theta}, q) = \mathrm{E}_q[\log p(\mathbf{X}|\mathbf{Z})] - KL(q(\mathbf{Z}|\mathbf{X}), p(\mathbf{Z}))$ with $KL(q(\mathbf{Z}|\mathbf{X}), p(\mathbf{Z})) \approx KL(p(\mathbf{Z}|\mathbf{X}), p(\mathbf{Z}))$, which has a similar form as eq. 10 above, but computes the divergence across trajectories (time), not across space).

**References**

1. Wilson HR. Spikes, decisions, and actions: the dynamical foundations of neurosciences: Oxford University Press; 1999.
2. Breakspear M. Dynamic models of large-scale brain activity. Nature Neuroscience. 2017;20:340. doi: 10.1038/nn.4497.
3. Izhikevich EM. Dynamical Systems in Neuroscience: MIT Press; 2007.
4. Hopfield JJ. Neural networks and physical systems with emergent collective computational abilities. Proceedings of the National Academy of Sciences U S A. 1982;79(8):2554-8. doi: 10.1073/pnas.79.8.2554.
5. Wang XJ. Synaptic reverberation underlying mnemonic persistent activity. Trends in neurosciences. 2001;24(8):455-63. Epub 2001/07/31. PubMed PMID: 11476885.
6. Durstewitz D, Seamans JK, Sejnowski TJ. Neurocomputational models of working memory. Nature Neuroscience. 2000;3 1184-91. Epub 2000/12/29. doi: 10.1038/81460. PubMed PMID: 11127836.
7. Albantakis L, Deco G. The encoding of alternatives in multiple-choice decision-making. BMC Neuroscience. 2009;10(1):166.
8. Rabinovich MI, Huerta R, Varona P, Afraimovich VS. Transient cognitive dynamics, metastability, and decision making. PLoS Computational Biology. 2008;4(5):e1000072.
9. Rabinovich M, Huerta R, Laurent G. Transient dynamics for neural processing. Science. 2008;321(5885):48-50.
10. Romo R, Brody CD, Hernández A, Lemus L. Neuronal correlates of parametric working memory in the prefrontal cortex. Nature. 1999;399(6735):470-3.
11. Machens CK, Romo R, Brody CD. Flexible control of mutual inhibition: a neural model of two-interval discrimination. Science. 2005;307(5712):1121-4.
12. Rabinovich MI, Varona P. Robust transient dynamics and brain functions. Frontiers in Computational Neuroscience. 2011;5:24-. doi: 10.3389/fncom.2011.00024. PubMed PMID: 21716642.
13. Seung HS, Lee DD, Reis BY, Tank DW. Stability of the memory of eye position in a recurrent network of conductance-based model neurons. Neuron. 2000;26(1):259-71.
14. Durstewitz D. Self-organizing neural integrator predicts interval times through climbing activity. Journal of Neuroscience. 2003;23(12):5342-53.
15. Balaguer-Ballester E, Moreno-Bote R, Deco G, Durstewitz D. Metastable dynamics of neural ensembles. Frontiers in Systems Neuroscience. 2017;11:99.
16. Smith AC, Brown EN. Estimating a state-space model from point process observations. Neural computation. 2003;15(5):965-91. Epub 2003/06/14. doi: 10.1162/089976603765202622. PubMed PMID: 12803953.
17. Paninski L, Ahmadian Y, Ferreira DG, Koyama S, Rahnama Rad K, Vidne M, et al. A new look at state-space models for neural data. J Comput Neurosci. 2010;29(1-2):107-26.
18. Ryali S, Supekar K, Chen T, Menon V. Multivariate dynamical systems models for estimating causal interactions in fMRI. NeuroImage. 2011;54(2):807-23. Epub 2010/10/05. doi: 10.1016/j.neuroimage.2010.09.052. PubMed PMID: 20884354; PubMed Central PMCID: PMCPmc2997172.
19. Macke JH, Buesing L, Sahani M. Estimating State and Parameters in State Space Models of Spike Trains. In: Chen Z, editor. Advanced State Space Methods for Neural and Clinical Data. Cambridge, UK: Cambridge University Press; 2015. p. 137-59.
20. Yu BM, Cunningham JP, Santhanam G, Ryu SI, Shenoy KV, Sahani M. Gaussian-process factor analysis for low-dimensional single-trial analysis of neural population activity. Journal of Neurophysiology. 2009;102(1):614-35. Epub 04/08. doi: 10.1152/jn.90941.2008. PubMed PMID: 19357332.
21. Friston KJ, Harrison L, Penny W. Dynamic causal modelling. NeuroImage. 2003;19(4):1273-302.
22. Balaguer-Ballester E, Lapish CC, Seamans JK, Durstewitz D. Attracting dynamics of frontal cortex ensembles during memory-guided decision-making. PLoS Computational Biology. 2011;7(5):e1002057.

23. Lapish CC, Balaguer-Ballester E, Seamans JK, Phillips aG, Durstewitz D. Amphetamine Exerts Dose-Dependent Changes in Prefrontal Cortex Attractor Dynamics during Working Memory. Journal of Neuroscience. 2015;35(28):10172-87.
24. Strogatz SH. Nonlinear dynamics and chaos: with applications to physics, biology, chemistry, and engineering: CRC Press; 2018.
25. Durstewitz D. Advanced Data Analysis in Neuroscience: Integrating statistical and computational models: Springer; 2017.
26. Funahashi K-i, Nakamura Y. Approximation of dynamical systems by continuous time recurrent neural networks. Neural Networks. 1993;6(6):801-6.
27. Kimura M, Nakano R. Learning dynamical systems by recurrent neural networks from orbits. Neural Networks. 1998;11(9):1589-99.
28. Trischler AP, D'Eleuterio GM. Synthesis of recurrent neural networks for dynamical system simulation. Neural Networks. 2016;80:67-78.
29. Yu BM, Afshar A, Santhanam G, Ryu S, Shenoy K, Sahani M, editors. Extracting dynamical structure embedded in neural activity. Advances in Neural Information Processing Systems 18; 2005: MIT Press.
30. Roweis S, Ghahramani Z. An EM algorithm for identification of nonlinear dynamical systems. 2000.
31. Durstewitz D. A state space approach for piecewise-linear recurrent neural networks for identifying computational dynamics from neural measurements. PLoS Computational Biology. 2017;13(6):e1005542. Epub 2017/06/03. doi: 10.1371/journal.pcbi.1005542. PubMed PMID: 28574992; PubMed Central PMCID: PMCPmc5456035.
32. Kingma DP, Welling M. Auto-encoding variational bayes. arXiv preprint arXiv:13126114. 2013.
33. Chung J, Kastner K, Dinh L, Goel K, Courville AC, Bengio Y, editors. A recurrent latent variable model for sequential data. Advances in neural information processing systems; 2015.
34. Bayer J, Osendorfer C. Learning stochastic recurrent networks. arXiv preprint arXiv:14117610v3. 2015.
35. Zhao Y, Park IM. Variational Joint Filtering. arXiv:170709049v3. 2018.
36. Pandarinath C, O'Shea DJ, Collins J, Jozefowicz R, Stavisky SD, Kao JC, et al. Inferring single-trial neural population dynamics using sequential auto-encoders. Nature methods. 2018;15(10):805-15. Epub 2018/09/19. doi: 10.1038/s41592-018-0109-9. PubMed PMID: 30224673.
37. Song HF, Yang GR, Wang X-J. Training excitatory-inhibitory recurrent neural networks for cognitive tasks: A simple and flexible framework. PLoS Computational Biology. 2016;12(2):e1004792.
38. Yang GR, Joglekar MR, Song HF, Newsome WT, Wang X-J. Task representations in neural networks trained to perform many cognitive tasks. Nature Neuroscience. 2019;22(2):297-306. doi: 10.1038/s41593-018-0310-2.
39. Hertäg L, Durstewitz D, Brunel N. Analytical approximations of the firing rate of an adaptive exponential integrate-and-fire neuron in the presence of synaptic noise. Frontiers in Computational Neuroscience. 2014;8:116.
40. Worsley KJ, Friston KJ. Analysis of fMRI time-series revisited—again. NeuroImage. 1995;2(3):173-81.
41. Durbin J, Koopman SJ. Time series analysis by state space methods: OUP Oxford; 2012.
42. LeCun Y, Bengio Y, Hinton G. Deep learning. Nature. 2015;521(7553):436-44. Epub 2015/05/29. doi: 10.1038/nature14539. PubMed PMID: 26017442.
43. Hinton GE, Osindero S, Teh YW. A fast learning algorithm for deep belief nets. Neural Comput. 2006;18(7):1527-54. Epub 2006/06/13. doi: 10.1162/neco.2006.18.7.1527. PubMed PMID: 16764513.
44. Goodfellow I, Bengio Y, Courville A, Bengio Y. Deep learning: MIT press Cambridge; 2016.
45. Talathi SS, Vartak A. Improving performance of recurrent neural network with relu nonlinearity. arXiv preprint arXiv:151103771. 2015.
46. Abarbanel HDI, Rozdeba PJ, Shirman S. Machine Learning: Deepest Learning as Statistical Data Assimilation Problems. Neural computation. 2018;30(8):2025-55. Epub 2018/06/13. doi: 10.1162/neco_a_01094. PubMed PMID: 29894650.
47. Lorenz EN. Deterministic nonperiodic flow. Journal of the Atmospheric Sciences. 1963;20(2):130-41.

48. Wood SN. Statistical inference for noisy nonlinear ecological dynamic systems. Nature. 2010;466(7310):1102.
49. Takens F. Detecting strange attractors in turbulence. In: Rand DA, Young L-S, editors. Dynamical Systems and Turbulence, Lecture notes in Mathematics. 898: Springer-Verlag; 1981. p. 366-81.
50. Sauer T, Yorke JA, Casdagli M. Embedology. Journal of Statistical Physics. 1991;65(3):579-616.
51. Kantz H, Schreiber T. Nonlinear time series analysis: Cambridge University Press; 2004.
52. van der Pol B. LXXXVIII. On "relaxation-oscillations". The London, Edinburgh, and Dublin Philosophical Magazine and Journal of Science. 1926;2(11):978-92. doi: 10.1080/14786442608564127.
53. Archer E, Park IM, Buesing L, Cunningham J, Paninski L. Black box variational inference for state space models. arXiv preprint arXiv:151107367. 2015.
54. Koppe G, Gruppe H, Sammer G, Gallhofer B, Kirsch P, Lis S. Temporal unpredictability of a stimulus sequence affects brain activation differently depending on cognitive task demands. NeuroImage. 2014;101:236-44. Epub 2014/07/16. doi: 10.1016/j.neuroimage.2014.07.008. PubMed PMID: 25019681.
55. Owen AM, McMillan KM, Laird AR, Bullmore E. N-back working memory paradigm: a meta-analysis of normative functional neuroimaging studies. Human brain mapping. 2005;25(1):46-59. Epub 2005/04/23. doi: 10.1002/hbm.20131. PubMed PMID: 15846822.
56. Tsuda I. Chaotic itinerancy and its roles in cognitive neurodynamics. Current Opinion in Neurobiology. 2015;31:67-71.
57. Wang X-J. Probabilistic decision making by slow reverberation in cortical circuits. Neuron. 2002;36(5):955-68.
58. Laurent G, Stopfer M, Friedrich RW, Rabinovich MI, Volkovskii A, Abarbanel HD. Odor encoding as an active, dynamical process: experiments, computation, and theory. Annual Review of Neuroscience. 2001;24(1):263-97.
59. Mante V, Sussillo D, Shenoy KV, Newsome WT. Context-dependent computation by recurrent dynamics in prefrontal cortex. Nature. 2013;503(7474):78-84. Epub 2013/11/10. doi: 10.1038/nature12742. PubMed PMID: 24201281; PubMed Central PMCID: PMCPmc4121670.
60. Churchland MM, Yu BM, Sahani M, Shenoy KV. Techniques for extracting single-trial activity patterns from large-scale neural recordings. Current opinion in neurobiology. 2007;17(5):609-18. doi: 10.1016/j.conb.2007.11.001. PubMed PMID: PMC2238690.
61. Nichols ALA, Eichler T, Latham R, Zimmer M. A global brain state underlies C. elegans sleep behavior. Science. 2017;356(6344):eaam6851. doi: 10.1126/science.aam6851.
62. Koiran P, Cosnard M, Garzon M. Computability with low-dimensional dynamical systems. Theoretical Computer Science. 1994;132(1-2):113-28.
63. Marr D. Vision: A computational investigation into the human representation and processing of visual information, henry holt and co. Inc, New York, NY. 1982;2(4.2).
64. Hertäg L, Hass J, Golovko T, Durstewitz D. An approximation to the adaptive exponential integrate-and-fire neuron model allows fast and predictive fitting to physiological data. Frontiers in Computational Neuroscience. 2012;6:62.
65. Fransén E, Tahvildari B, Egorov AV, Hasselmo ME, Alonso AA. Mechanism of graded persistent cellular activity of entorhinal cortex layer v neurons. Neuron. 2006;49(5):735-46.
66. Ozaki T. Time series modeling of neuroscience data: CRC Press; 2012.
67. Pathak J, Lu Z, Hunt BR, Girvan M, Ott E. Using machine learning to replicate chaotic attractors and calculate Lyapunov exponents from data. Chaos: An Interdisciplinary Journal of Nonlinear Science. 2017;27(12):121102.
68. Brunton SL, Proctor JL, Kutz JN. Discovering governing equations from data by sparse identification of nonlinear dynamical systems. Proceedings of the National Academy of Sciences U S A. 2016;113(15):3932-7.
69. Collins FS, Varmus H. A new initiative on precision medicine. The New England journal of medicine. 2015;372(9):793-5. Epub 2015/01/31. doi: 10.1056/NEJMp1500523. PubMed PMID: 25635347; PubMed Central PMCID: PMCPmc5101938.
70. Durstewitz D, Huys QJM, Koppe G. Psychiatric Illnesses as Disorders of Network Dynamics. arXiv:180906303. 2018.

71. Durstewitz D, Seamans JK. The dual-state theory of prefrontal cortex dopamine function with relevance to catechol-o-methyltransferase genotypes and schizophrenia. Biological Psychiatry. 2008;64(9):739-49.
72. Armbruster DJ, Ueltzhöffer K, Basten U, Fiebach CJ. Prefrontal cortical mechanisms underlying individual differences in cognitive flexibility and stability. Journal of Cognitive Neuroscience. 2012;24(12):2385-99.
73. Li X, Zhu D, Jiang X, Jin C, Zhang X, Guo L, et al. Dynamic functional connectomics signatures for characterization and differentiation of PTSD patients. Human brain mapping. 2014;35(4):1761-78. Epub 2013/05/15. doi: 10.1002/hbm.22290. PubMed PMID: 23671011; PubMed Central PMCID: PMCPmc3928235.
74. Damaraju E, Allen EA, Belger A, Ford JM, McEwen S, Mathalon DH, et al. Dynamic functional connectivity analysis reveals transient states of dysconnectivity in schizophrenia. NeuroImage Clinical. 2014;5:298-308. Epub 2014/08/28. doi: 10.1016/j.nicl.2014.07.003. PubMed PMID: 25161896; PubMed Central PMCID: PMCPmc4141977.
75. Rashid B, Damaraju E, Pearlson GD, Calhoun VD. Dynamic connectivity states estimated from resting fMRI Identify differences among Schizophrenia, bipolar disorder, and healthy control subjects. Frontiers in Human Neuroscience. 2014;8(897). doi: 10.3389/fnhum.2014.00897.
76. Smetters D, Majewska A, Yuste R. Detecting action potentials in neuronal populations with calcium imaging. Methods. 1999;18(2):215-21.
77. Shoham D, Glaser DE, Arieli A, Kenet T, Wijnbergen C, Toledo Y, et al. Imaging cortical dynamics at high spatial and temporal resolution with novel blue voltage-sensitive dyes. Neuron. 1999;24(4):791-802.
78. Koppe G, Guloksuz S, Reininghaus U, Durstewitz D. Recurrent Neural Networks in Mobile Sampling and Intervention. Schizophrenia bulletin. 2019;45(2):272-6. Epub 2018/11/30. doi: 10.1093/schbul/sby171. PubMed PMID: 30496527; PubMed Central PMCID: PMCPmc6403085.
79. Sugihara G, May R, Ye H, Hsieh C-h, Deyle E, Fogarty M, et al. Detecting Causality in Complex Ecosystems. Science. 2012;338(6106):496. doi: 10.1126/science.1227079.
80. Hershey JR, Olsen PA, editors. Approximating the Kullback Leibler divergence between Gaussian mixture models. Acoustics, Speech and Signal Processing, 2007 ICASSP 2007 IEEE International Conference on; 2007: IEEE.
81. Krzanowski W. Principles of multivariate analysis: OUP Oxford; 2000.
82. Dempster AP, Laird NM, Rubin DB. Maximum likelihood from incomplete data via the EM algorithm. Journal of the Royal Statistical Society Series B (methodological). 1977:1-38.
83. Kalman RE. A New Approach to Linear Filtering and Prediction Problems. Transactions of the ASME – Journal of Basic Engineering. 1960;(82 (Series D)):35-45. doi: citeulike-article-id:347166.
84. Rauch HE, Striebel CT, Tung F. Maximum likelihood estimates of linear dynamic systems. 1965;3(8):1445-50. doi: 10.2514/3.3166. PubMed PMID: pub.1015705383.
85. Koyama S, Pérez-Bolde LC, Shalizi CR, Kass RE. Approximate Methods for State-Space Models. Journal of the American Statistical Association. 2010;105(489):170-80. doi: 10.1198/jasa.2009.tm08326. PubMed PMID: PMC3132892.
86. Brugnano L, Casulli V. Iterative Solution of Piecewise Linear Systems. SIAM Journal on Scientific Computing. 2008;30(1):463-72. doi: 10.1137/070681867.
87. Rezende DJ, Mohamed S, Wierstra D. Stochastic backpropagation and approximate inference in deep generative models. arXiv preprint arXiv:14014082. 2014.
88. Manning CD, Raghavan P, Schütze M. Introduction to Information Retrieval: Cambridge University Press; 2008.
89. Gevins AS, Bressler SL, Cutillo BA, Illes J, Miller JC, Stern J, et al. Effects of prolonged mental work on functional brain topography. Electroencephalography and Clinical Neurophysiology. 1990;76(4):339-50. doi: 10.1016/0013-4694(90)90035-I.
90. Bengtsson T, Bickel P, Li B. Curse-of-dimensionality revisited: Collapse of the particle filter in very large scale systems. Probability and statistics: Essays in honor of David A Freedman: Institute of Mathematical Statistics; 2008. p. 316-34.

91. Li T, Sun S, Sattar TP, Corchado JM. Fight sample degeneracy and impoverishment in particle filters: A review of intelligent approaches. Expert Systems with Applications. 2014;41(8):3944-54.

**Supplement**

**S1 Text. Model specification and inference.**

**PLRNN-BOLD-SSM model inference.** In the EM algorithm, we first aim to determine the posterior distribution $p(\mathbf{Z}|\mathbf{X})$ (E-Step), and – given this – then maximize the expectation of the joint ('complete data') log-likelihood $\mathrm{E}_{q(\mathbf{Z}|\mathbf{X})}[\log p(\mathbf{Z}, \mathbf{X}|\boldsymbol{\theta})] := Q(\boldsymbol{\theta}, \mathbf{Z})$ w.r.t. the parameters (M-Step). With the Gaussian noise assumptions (see eqns. 1-3, main manuscript), the expected joint log-likelihood is given by

(1) $Q(\boldsymbol{\Theta}, \mathbf{Z}) = -\frac{1}{2}\mathrm{E}_q[(\mathbf{z}_1 - \boldsymbol{\mu}_0 - \mathbf{C}\mathbf{s}_1)^\mathrm{T}\boldsymbol{\Sigma}^{-1}(\mathbf{z}_1 - \boldsymbol{\mu}_0 - \mathbf{C}\mathbf{s}_1)]$

$\quad -\frac{1}{2}\mathrm{E}_q[\sum_{t=2}^{T}(\mathbf{z}_t - \mathbf{A}\mathbf{z}_{t-1} - \mathbf{W}\varphi(\mathbf{z}_{t-1}) - \mathbf{h} - \mathbf{C}\mathbf{s}_t)^\mathrm{T}\boldsymbol{\Sigma}^{-1}(\mathbf{z}_t - \mathbf{A}\mathbf{z}_{t-1} - \mathbf{W}\varphi(\mathbf{z}_{t-1}) - \mathbf{h} - \mathbf{C}\mathbf{s}_t)]$

$\quad -\frac{1}{2}\mathrm{E}_q[\sum_{t=1}^{T}(\mathbf{x}_t - \mathbf{B}(hrf * \mathbf{z}_{\tau:t}) - \mathbf{J}\mathbf{r}_t)^\mathrm{T}\boldsymbol{\Gamma}^{-1}(\mathbf{x}_t - \mathbf{B}(hrf * \mathbf{z}_{\tau:t}) - \mathbf{J}\mathbf{r}_t)] - \frac{T}{2}(\log|\boldsymbol{\Sigma}| + \log|\boldsymbol{\Gamma}|)$,

where $\varphi(\mathbf{z}_t) := \max(\mathbf{z}_t, 0)$ is an element-wise piecewise linear (ReLU) activation function.

The convolution with the hemodynamic response function (HRF) spells out as $hrf * \mathbf{z}_{\tau:t} = \sum_{\tau=t-\Delta t}^{t} h_{t-\tau+1} \mathbf{z}_\tau$, where $h_{t-\tau+1}$ indexes the individual components of the HRF vector, and $\Delta t = 0 \dots n\text{-}1$ depends on the temporal resolution of the time series, i.e. on the length n of the HRF vector.

**State estimation (E-Step).** We assume that $p(\mathbf{Z}|\mathbf{X})$, like a Gaussian, could be specified by its first two moments, i.e. the mean and the covariance, and that, as for a Gaussian, the MAP estimator is a reasonably good approximation to the mean. Thus we aim to maximize the log-joint distribution over $\mathbf{X}$ and $\mathbf{Z}$, $\log p(\mathbf{Z}, \mathbf{X}|\boldsymbol{\theta})$ (as defined by $Q_\Omega^*(\mathbf{Z})$ in the main manuscript) w.r.t. $\mathbf{Z}$ [see also 17, 85].

We have reformulated the optimization criterion in eq. 6 (main manuscript) in 'big-matrix form', defining the set of 'active' states (i.e., $\mathbf{z}_t > 0$) through the binary vector $\mathbf{d}_\Omega := \mathrm{I}(\mathbf{z} > 0)$. The matrix $\mathbf{H} \in \mathbb{R}^{MT \times MT}$ in eq. 6 is a convolution matrix which for $M=1$ contains the elements of the HRF in the following form

$$\mathbf{H} = \begin{bmatrix} h_n & 0 & 0 & 0 & 0 & 0 & 0 & 0 & 0 \\ h_{n-1} & h_n & 0 & 0 & 0 & 0 & 0 & 0 & 0 \\ h_{n-2} & h_{n-1} & h_n & 0 & 0 & 0 & 0 & 0 & 0 \\ \cdots & h_{n-2} & h_{n-1} & h_n & 0 & 0 & 0 & 0 & 0 \\ h_1 & \cdots & h_{n-2} & h_{n-1} & h_n & 0 & 0 & 0 & 0 \\ 0 & h_1 & \cdots & h_{n-2} & h_{n-1} & h_n & 0 & 0 & 0 \\ 0 & 0 & h_1 & \cdots & h_{n-2} & h_{n-1} & h_n & 0 & 0 \\ 0 & 0 & 0 & h_1 & \cdots & h_{n-2} & h_{n-1} & h_n & 0 \\ 0 & 0 & 0 & 0 & h_1 & \cdots & h_{n-2} & h_{n-1} & h_n \end{bmatrix}$$

where $h_i$, $i=1\dots n$, denote the HRF vector elements.

For $M>1$, we would add additional columns by inserting $M$-1 zeros between each element of $\mathbf{H}$, and add additional rows by duplicating each row $M$-1 times, and shifting it by 1…$M$-1 positions, respectively.

Below we will also give the full structure of the block-banded matrices $\in \mathbb{R}^{MT \times MT}$ that occur in eq. 6, restated here for convenience:

$$Q_\Omega^*(\mathbf{Z}) = -\frac{1}{2}[\mathbf{z}^T(\mathbf{U}_0 + \mathbf{D}_\Omega^T\mathbf{U}_1 + \mathbf{U}_1^T\mathbf{D}_\Omega + \mathbf{D}_\Omega^T\mathbf{U}_2\mathbf{D}_\Omega + \mathbf{H}^T\mathbf{U}_3\mathbf{H})\mathbf{z} - \mathbf{z}^T(\mathbf{v}_0 + \mathbf{d}_\Omega^T \circ \mathbf{v}_1 + \mathbf{H}^T\mathbf{v}_2)$$
$$- (\mathbf{v}_0 + \mathbf{d}_\Omega^T \circ \mathbf{v}_1 + \mathbf{H}^T\mathbf{v}_2)^T\mathbf{z}] + \text{const}$$

Here, $\circ$ denotes the Hadamard product, all terms that do not depend on $\mathbf{z}$ are collected in $\text{const}$, and the matrices and vectors are defined as follows:

$$\mathbf{U}_0 = \begin{bmatrix} \Sigma^{-1} + \mathbf{A}^T\Sigma^{-1}\mathbf{A} & -\mathbf{A}^T\Sigma^{-1} & 0 & \cdots & 0 \\ -\Sigma^{-1}\mathbf{A} & \ddots & \ddots & \ddots & \vdots \\ 0 & \ddots & \ddots & \ddots & 0 \\ \vdots & \ddots & \ddots & \Sigma^{-1} + \mathbf{A}^T\Sigma^{-1}\mathbf{A} & -\mathbf{A}^T\Sigma^{-1} \\ 0 & \cdots & 0 & -\Sigma^{-1}\mathbf{A} & \Sigma^{-1} \end{bmatrix},$$

$$\mathbf{U}_1 = \begin{bmatrix} \mathbf{W}^T\Sigma^{-1}\mathbf{A} & -\mathbf{W}^T\Sigma^{-1} & 0 & \cdots & 0 \\ 0 & \ddots & \ddots & \ddots & \vdots \\ \vdots & \ddots & \mathbf{W}^T\Sigma^{-1}\mathbf{A} & -\mathbf{W}^T\Sigma^{-1} & 0 \\ \vdots & \ddots & 0 & \mathbf{W}^T\Sigma^{-1}\mathbf{A} & -\mathbf{W}^T\Sigma^{-1} \\ 0 & 0 & \cdots & 0 & 0 \end{bmatrix},$$

$$\mathbf{U}_2 = \begin{bmatrix} \mathbf{W}^T\Sigma^{-1}\mathbf{W} & 0 & 0 & \cdots & 0 \\ 0 & \ddots & \ddots & \ddots & \vdots \\ \vdots & \ddots & \mathbf{W}^T\Sigma^{-1}\mathbf{W} & \ddots & 0 \\ \vdots & \ddots & 0 & \mathbf{W}^T\Sigma^{-1}\mathbf{W} & 0 \\ 0 & 0 & \cdots & 0 & 0 \end{bmatrix},$$

$$\mathbf{U}_3 = \begin{bmatrix} \mathbf{B}^T\Gamma^{-1}\mathbf{B} & 0 & 0 & \cdots & 0 \\ 0 & \ddots & \ddots & \ddots & \vdots \\ \vdots & \ddots & \ddots & \ddots & 0 \\ \vdots & \ddots & 0 & \mathbf{B}^T\Gamma^{-1}\mathbf{B} & 0 \\ 0 & 0 & \cdots & 0 & \mathbf{B}^T\Gamma^{-1}\mathbf{B} \end{bmatrix},$$

$$\mathbf{v}_0 = \begin{bmatrix} \Sigma^{-1}\mathbf{Cs}_1 - \mathbf{A}^T\Sigma^{-1}(\mathbf{Cs}_2 + \boldsymbol{\theta}) + \Sigma^{-1}\boldsymbol{\mu}_0 \\ \vdots \\ \Sigma^{-1}(\mathbf{Cs}_t + \boldsymbol{\theta}) - \mathbf{A}^T\Sigma^{-1}(\mathbf{Cs}_{t+1} + \boldsymbol{\theta}) \\ \vdots \\ \Sigma^{-1}(\mathbf{Cs}_T + \boldsymbol{\theta}) \end{bmatrix},$$

$$\mathbf{v}_1 = \begin{bmatrix} -\mathbf{W}^T\Sigma^{-1}(\mathbf{Cs}_2 + \boldsymbol{\theta}) \\ -\mathbf{W}^T\Sigma^{-1}(\mathbf{Cs}_{t+1} + \boldsymbol{\theta}) \\ \vdots \\ -\mathbf{W}^T\Sigma^{-1}(\mathbf{Cs}_T + \boldsymbol{\theta}) \\ 0 \end{bmatrix},$$

$$\mathbf{v}_2 = \begin{bmatrix} \mathbf{B}^T\Gamma^{-1}\mathbf{x}_1 - \mathbf{B}^T\Gamma^{-1}\mathbf{Jr}_1 \\ \vdots \\ \mathbf{B}^T\Gamma^{-1}\mathbf{x}_t - \mathbf{B}^T\Gamma^{-1}\mathbf{Jr}_t \\ \vdots \\ \mathbf{B}^T\Gamma^{-1}\mathbf{x}_T - \mathbf{B}^T\Gamma^{-1}\mathbf{Jr}_T \end{bmatrix}.$$

**Parameter estimation (M-Step).** In the M-Step, we maximize $Q(\boldsymbol{\theta}, \mathbf{Z})$ given the state expectations returned by the E-Step w.r.t. to the parameters, which can be done analytically. The closed-form solution for the observation model parameters was given in the main manuscript; here we add the solutions for the latent model parameters. Specifically, we solve for $\mathbf{A}$, $\mathbf{W}$, $\mathbf{h}$, and $\mathbf{C}$ simultaneously by arranging these parameters horizontally within a matrix $\mathbf{L} := [\mathbf{A}\ \mathbf{W}\ \mathbf{h}\ \mathbf{C}] \in \mathbb{R}^{M\times(2M+1+K)}$, and defining the vector of predictor variables as $\mathbf{o}_t := [\mathbf{z}_{t-1},\ \varphi(\mathbf{z}_{t-1}), 1, \mathbf{s}_t] \in \mathbb{R}^{(2M+1+K)\times 1}$.

Since matrices **A** and **W** are not full (but contain to-be-estimated parameters only along the diagonal or off the diagonal, respectively), we solve for each row j of **L** in turn as

$$(2)\quad \mathbf{L}_{j,k(j)} = \left(\sum_{t=2}^{T} \mathrm{E}[z_{j,t}\mathbf{o}_{k(j),t}^{\mathrm{T}}]\right)\left(\sum_{t=2}^{T} \mathrm{E}[\mathbf{o}_{k(j),t}\mathbf{o}_{k(j),t}^{\mathrm{T}}]\right)^{-1}$$

where *k(j)* is a set of row indices which pick out those rows in $\mathbf{o}_t$ that correspond to those parameters actually defined, i.e. which squeeze out all zeros from the *j*th row of **L**, and accordingly from the respective rows/columns on the r.h.s. of eq. 2.

Let us define the following expectation sums:

$\mathbf{E}_1 := \sum_{t=2}^{T} \mathrm{E}[\varphi(\mathbf{z}_{t-1})\varphi(\mathbf{z}_{t-1})^{\mathrm{T}}],\quad \mathbf{E}_2 := \sum_{t=2}^{T} \mathrm{E}[\mathbf{z}_t \mathbf{z}_{t-1}^{\mathrm{T}}],\quad \mathbf{E}_3 := \sum_{t=2}^{T} \mathrm{E}[\mathbf{z}_{t-1}\mathbf{z}_{t-1}^{\mathrm{T}}],$

$\mathbf{E}_4 := \sum_{t=2}^{T} \mathrm{E}[\varphi(\mathbf{z}_{t-1})\mathbf{z}_{t-1}^{\mathrm{T}}],\quad \mathbf{E}_5 := \sum_{t=2}^{T} \mathrm{E}[\mathbf{z}_t \varphi(\mathbf{z}_{t-1})^{\mathrm{T}}],\quad \mathbf{F}_3 := \sum_{t=2}^{T} \mathbf{s}_t \mathrm{E}[\mathbf{z}_{t-1}^{\mathrm{T}}],$

$\mathbf{F}_4 := \sum_{t=2}^{T} \mathbf{s}_t \mathrm{E}[\varphi(\mathbf{z}_{t-1})^{\mathrm{T}}],\quad \mathbf{F}_5 := \sum_{t=2}^{T} \mathrm{E}[\mathbf{z}_t]\mathbf{s}_t^{\mathrm{T}},\quad \mathbf{F}_6 := \sum_{t=2}^{T} \mathbf{s}_t \mathbf{s}_t^{\mathrm{T}},$

$\mathbf{G}_{1\Delta} := \sum_{t=1+\Delta}^{T-1+\Delta} \mathrm{E}[\mathbf{z}_t],\quad \mathbf{G}_2 := \sum_{t=2}^{T} \mathbf{s}_t^{\mathrm{T}},\quad \mathbf{G}_3 := \sum_{t=2}^{T} \mathrm{E}[\varphi(\mathbf{z}_{t-1})].$

With this, the vector and matrix in eq. 2 can be written as

$$(3)\quad \sum_{t=2}^{T} \mathrm{E}[z_{j,t}\mathbf{o}_t^{\mathrm{T}}] = [\mathbf{E}_{2,j}\ \mathbf{E}_{5,j}\ \mathbf{G}_{11,j}\ \mathbf{F}_{5,j}],$$

and

$$(4)\quad \sum_{t=2}^{T} \mathrm{E}[\mathbf{o}_t \mathbf{o}_t^{\mathrm{T}}] = \begin{bmatrix} \mathbf{E}_3 & \mathbf{E}_4^{\mathrm{T}} & \mathbf{G}_{10} & \mathbf{F}_3^{\mathrm{T}} \\ \mathbf{E}_4 & \mathbf{E}_1 & \mathbf{G}_3 & \mathbf{F}_4^{\mathrm{T}} \\ \mathbf{G}_{10}^{\mathrm{T}} & \mathbf{G}_3^{\mathrm{T}} & T-1 & \mathbf{G}_2 \\ \mathbf{F}_3 & \mathbf{F}_4 & \mathbf{G}_2^{\mathrm{T}} & \mathbf{F}_6 \end{bmatrix},$$

from which we select the columns *k(j)* from eq. 3, and the rows and columns *k(j)* out of eq. 4 to produce the solution for the *j*-th row in eq. 2.

Finally, the estimate for the initial condition is given by

$$\boldsymbol{\mu}_0 = \mathrm{E}[\mathbf{z}_1] - \mathbf{C}\mathbf{s}_1.$$

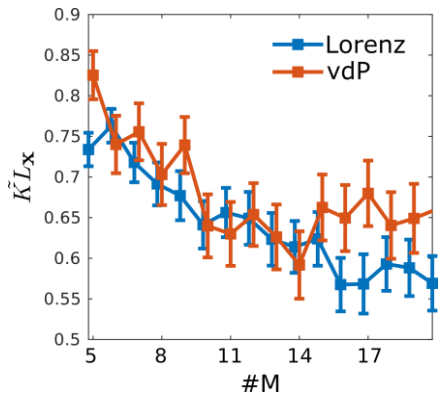

**S1 Fig. Dependence of $\widetilde{KL}_x$ on number of latent states (*M*) for the vdP (red) and Lorenz (blue) systems.** *M*=14 seems to be about optimal for vdP, while *M*≈16 may be about optimal for the Lorenz system.

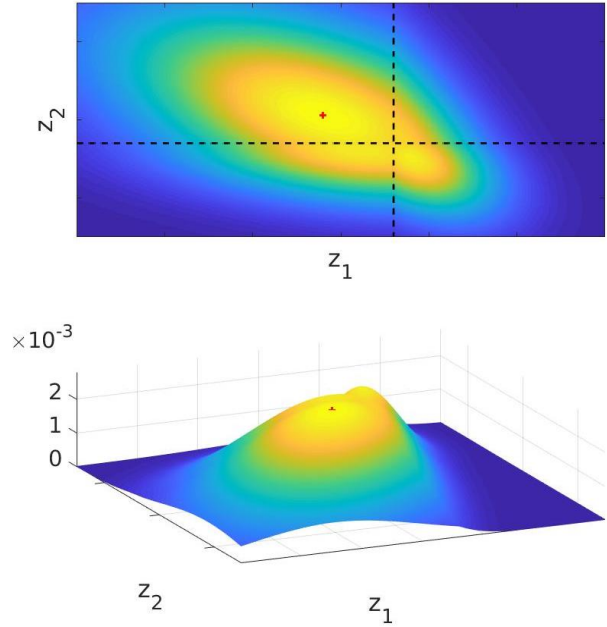

**S2 Fig. Likelihood landscape.** Illustration of the model's likelihood landscape as a function of a single latent state across two consecutive time steps, $z_1$ and $z_2$. The joint likelihood $p(\mathbf{X},\mathbf{Z})$ consists of piecewise Gaussians which cut off at the zeros of the states; often they will cluster near the origin and give rise to a strongly elevated plateau of high-likelihood solutions, close to one full Gaussian. Red cross indicates MAP estimate.

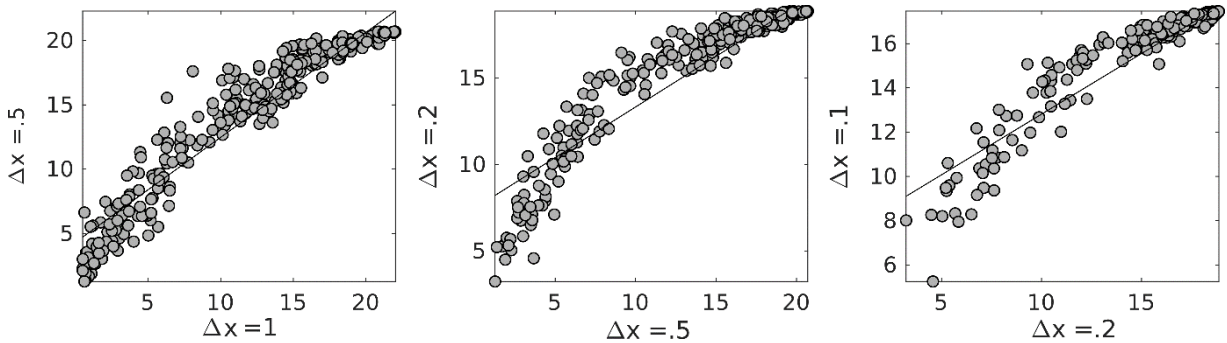

**S3 Fig. Agreement in Kullback Leibler divergence $KL_\mathbf{x}$ (eq. 9) on discretized observation space for different bin sizes (assessed for the Lorenz system).** A. $KL_\mathbf{x}$ for bin size $\Delta\mathbf{x}$=1 (x-axis) against bin size $\Delta\mathbf{x}$=.5 (y-axis). B. Same as A for bin size $\Delta\mathbf{x}$=.5 (x-axis) against $\Delta\mathbf{x}$=.2 (y-axis). C. Same as A. for bin size $\Delta\mathbf{x}$=.2 (x-axis) against $\Delta\mathbf{x}$=.1 (y-axis). Measures at different bin sizes are nearly monotonically related such that rank information on the quality of DS retrieval is conserved. However, the $KL_\mathbf{x}$ spread is largest for $\Delta\mathbf{x}$=1 such that qualitative differences in DS retrieval are differentiated more easily for this bin size, and hence this bin size was chosen for the evaluation in the main manuscript.